\documentclass{article} 
\usepackage{iclr2025_conference,times}

\usepackage{amsmath,amsfonts,bm}

\def\eqref#1{equation~\ref{#1}}

\def\1{\bm{1}}

\DeclareMathAlphabet{\mathsfit}{\encodingdefault}{\sfdefault}{m}{sl}
\SetMathAlphabet{\mathsfit}{bold}{\encodingdefault}{\sfdefault}{bx}{n}

\usepackage{hyperref}
\usepackage{graphicx}
\usepackage{subcaption}
\usepackage{url}
\usepackage{multirow}
\usepackage{wrapfig}
\usepackage{enumitem}
\usepackage{algorithm}
\usepackage{algorithmic}
\usepackage{amsmath}

\title{YOSO: You-Only-Sample-Once via Compressed Sensing for Graph Neural Network Training}

\author{Yi Li$^1$, ~Zhichun Guo$^{2, 3}$, ~Guanpeng Li$^4$, ~Bingzhe Li$^1$ \\
$^1$ The University of Texas at Dallas\\
$^{2}$ University of Washington,~$^{3}$ University of Notre Dame, ~$^4$ University of Iowa\\
\texttt{$^1$\{Yi.Li3, Bingzhe.Li\}@utdallas.edu}, ~\texttt{$^{2, 3}$\{Zguo5\}@nd.edu}\\ 
\texttt{$^4$\{guanpeng-li\}@uiowa.edu}
}

\iclrfinalcopy 
\begin{document}

\maketitle

\begin{abstract}
Graph neural networks (GNNs) have become essential tools for analyzing non-Euclidean data across various domains. During training stage, sampling plays an important role in reducing latency by limiting the number of nodes processed, particularly in large-scale applications. However, as the demand for better prediction performance grows, existing sampling algorithms become increasingly complex, leading to significant overhead. To mitigate this, we propose YOSO (You-Only-Sample-Once), an algorithm designed to achieve efficient training while preserving prediction accuracy. YOSO introduces a compressed sensing (CS)-based sampling and reconstruction framework, where nodes are sampled once at input layer, followed by a lossless reconstruction at the output layer per epoch. By integrating the reconstruction process with the loss function of specific learning tasks, YOSO not only avoids costly computations in traditional compressed sensing (CS) methods, such as orthonormal basis calculations, but also ensures high-probability accuracy retention which equivalent to full node participation. Experimental results on node classification and link prediction demonstrate the effectiveness and efficiency of YOSO, reducing GNN training by an average of 75\% compared to state-of-the-art methods, while maintaining accuracy on par with top-performing baselines.
\end{abstract}

\section{Introduction}\label{sec:introduction}
Graph Neural Networks (GNNs)~\citep{kipf2016semi, hamilton2017inductive, velivckovic2017graph, chen2018fastgcn, chiang2019cluster, zou2019layer} have become pivotal in analyzing graph data across various domains, such as social network~\citep{guo2020deep}, protein interactions~\citep{reau2023deeprank}, and transportation systems~\citep{liu2021community}. As graphs rapidly grow, long training time becomes a crucial factor impeding the wide utilization of GNNs in real world. To mitigate this issue, various sampling strategies such as node-wise~\citep{hamilton2017inductive, chen2017stochastic}, layer-wise~\citep{chen2018fastgcn, zou2019layer, huang2018adaptive}, and subgraph-based methods~\citep{chiang2019cluster, zeng2019graphsaint} have been developed. These sampling strategies reduce the amount of data that required to sustain training and potentially shorten the training time. However, with the increasing complexity of sampling algorithms, GNNs have struggled to maintain training efficiency in large-scale applications, such as IGB dataset~\citep{khatua2023igb}.

\begin{figure}[!t]
\begin{minipage}{0.5\linewidth}
    \centerline{\includegraphics[width=\linewidth]{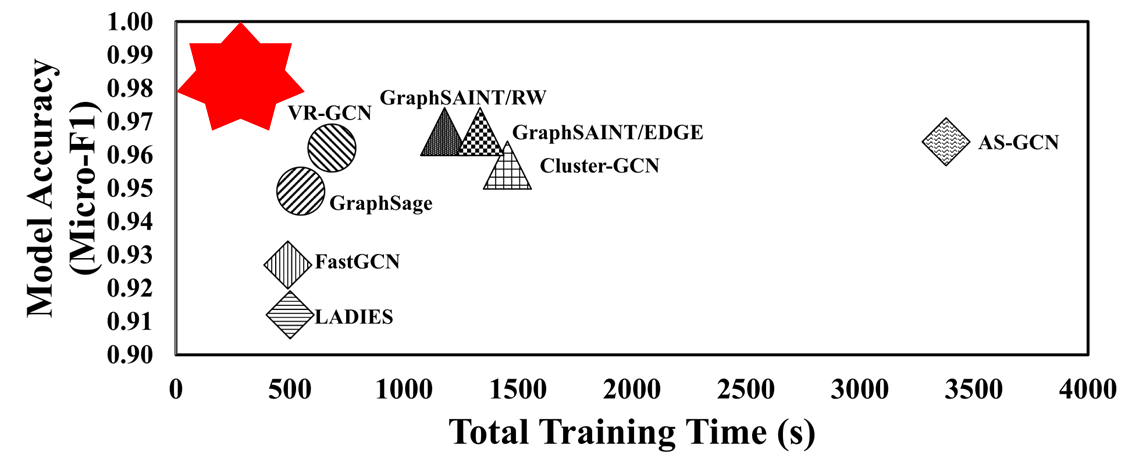}}
    \subcaption{Total training time v.s. model accuracy}
\end{minipage}
\begin{minipage}{0.5\linewidth}
    \centerline{\includegraphics[width=\linewidth]{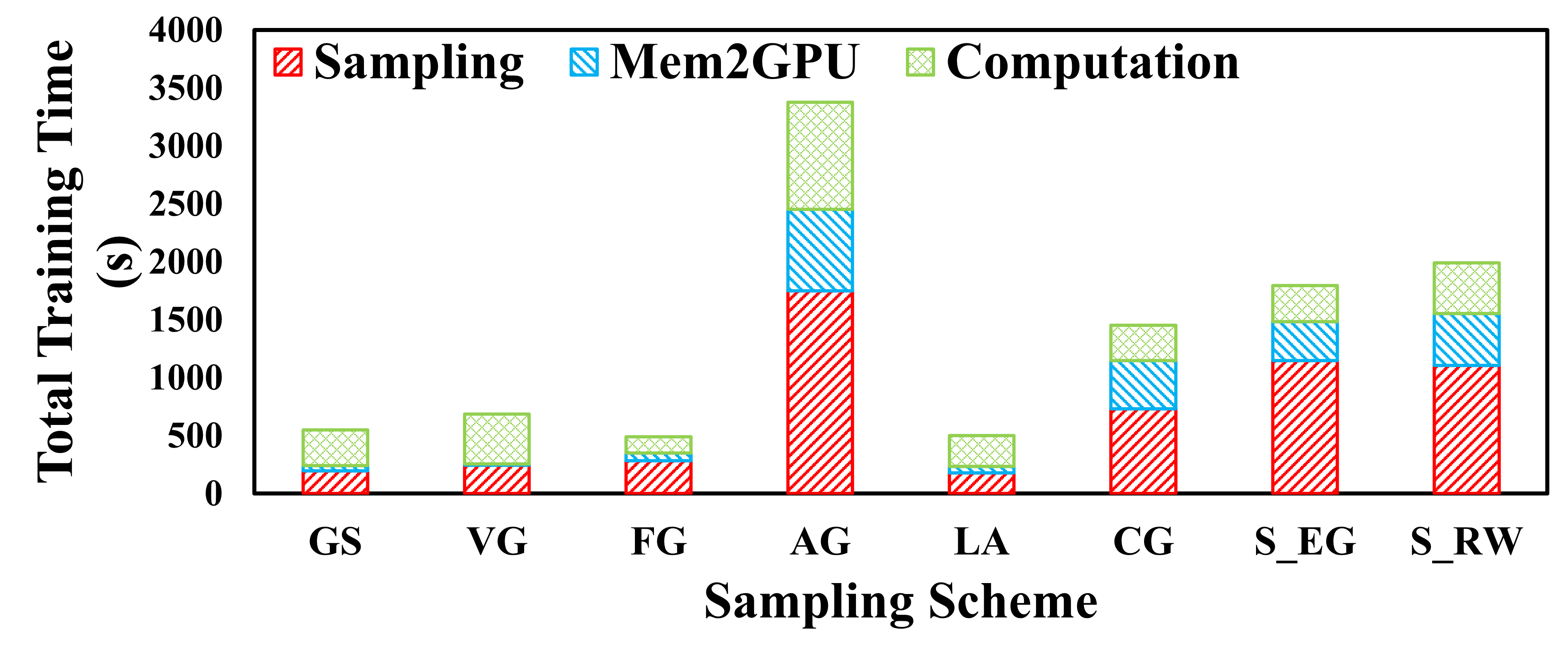}}
    \subcaption{Total training time breakdown on Reddit}
\end{minipage}

\caption{Total training time (with breakdown) and model accuracy for different sampling schemes, including GS (GraphSage~\citep{hamilton2017inductive}), VG (VR-GCN~\citep{chen2017stochastic}), FG (FastGCN~\citep{chen2018fastgcn}), AG (AS-GCN~\citep{huang2018adaptive}), LA (LADIES~\citep{zou2019layer}), CG (Cluster-GCN~\citep{chiang2019cluster}) and two versions of GraphSAINT~\citep{zeng2019graphsaint}: S\_EG (EDGE) and S\_RW (Random Walk), on Reddit dataset~\citep{hamilton2017inductive}. The seven-pointed red star marks the contribution of this paper.}
\vspace{-5mm}
\label{fig:result_motivation}
\end{figure}

Theoretically, the model accuracy loss caused by sampling algorithms stems from the bias and variance introduced by estimating the overall data based on the samples~\citep{huang2018adaptive}. Unlike the unbiased and variance-free GCN~\citep{kipf2016semi, huang2018adaptive} that utilize all training nodes, low time complexity sampling methods struggle to accurately estimate both graph structure and embeddings~\citep{jin2020graph}, potentially degrading the outcomes. As a result, recent sampling algorithms focus solely on improving accuracy while overlooking high computational cost, have become increasingly complex. This highlights a significant gap as shown in Figure~\ref{fig:result_motivation}(a): finding a method that achieves both high accuracy and efficiency.

To reveal the large overhead introduced by sampling in GNN training, we conduct empirical evaluations for state-of-the-art (SOTA) sampling schemes with Reddit dataset~\citep{hamilton2017inductive}. As shown in Figure~\ref{fig:result_motivation}(b), we break down the total training time into three non-overlapping components: (1) Sampling, (2) Mem2GPU: refers to transferring data to GPU memory, and (3) Computation: all processes on GPU. Our results indicate that sampling stage occupies 35.7\% to 64\% of the total training time across various sampling algorithms, making it a significant overhead when considering both training efficiency and model accuracy. For instance, as a representative of layer-wise, AS-GCN~\citep{huang2018adaptive} spends 55.6\% of the total training time on sampling but only achieves suboptimal accuracy. Subgraph-based sampling methods, although achieving the highest model accuracy, incur the most significant overhead with sampling stage accounting for up to 64\% of the total training time. Node-wise sampling falls between these two paradigms in terms of overall performance. For example, VR-GCN spends 685.72 seconds on sampling and achieves a Micro-F1 score of 0.962. 

To fill identified research gap, we propose YOSO (You-Only-Sample-Once), a novel approach applies compressed sensing (CS)~\citep{candes2006near} technique to GNN sampling. YOSO reimagines the feature matrix as multi-channel signal and utilize adapted CS to reduce the amount of computation involved in the training by transferring the feature matrix to another domain with high sparsity. YOSO enables training with only $M$ nodes sampled from the graph with $N$ nodes where $M \ll N$, followed by a nearly lossless reconstruction which guarantees the model accuracy closely aligns with zero bias and variance as if all $N$ nodes were used for training. Moreover, sampling in YOSO is designed to occur only once at the beginning of the training. This involves determining the sampling matrix, represented as $\mathbf{\Phi}$, based on the characteristics of the graph dataset. Subsequently, the reconstruction process takes place after each forward propagation, achieved by integrating
the reconstruction process with the loss function of specific learning tasks, i.e., cross-entropy loss in node classification, to guide the backward propagation. Thus, YOSO streamlines the entire training process by eliminating the need for continuous resampling through entire training and ensures every step of learning is informed by an optimally reconstructed data state, significantly enhancing both the efficiency and efficacy of the model training. We summarize our contributions below.
\begin{itemize}[topsep=0pt,itemsep=-1ex,partopsep=1ex,parsep=1ex, leftmargin=*]
    \item We propose a novel sampling method called YOSO, which significantly reduces GNN training time by performing one-time sampling for the entire training while maintaining strong prediction accuracy through a nearly lossless reconstruction of the embedding matrix.
    \item YOSO eliminates the need for expensive computations typically associated with combining CS with GNN sampling, thereby making the sampling process highly efficient.
    \item Experimental results demonstrate the effectiveness of YOSO on both node classification and link prediction tasks. Specifically, YOSO significantly reduces overall training time by an average of around 75\%  while preserving model accuracy. Ablation studies further reveal that YOSO achieves near-zero bias and variance, effectively reconstructing the embedding matrix with minimal error.
\end{itemize}

\section{Preliminaries}\label{sec: pre}
\textbf{Graph Neural Networks (GNNs)} operate on a graph, represented as \( G = \{ V, E, \hat{\mathbf{A}}, \mathbf{X} \} \), where \( V = \{ 1, 2, \dots, N \} \) represents the set of nodes, \( E = \{ (i, j) \mid i, j \in V \} \) defines the edges, and \( \hat{\mathbf{A}} \in \mathbb{R}^{N \times N} \) is a matrix that encoding the connection properties between nodes, i.e., adjacency matrix or normalized Laplacian matrix. \( \mathbf{X} \in \mathbb{R}^{N \times d} \) is feature matrix, where \( d \) is the feature dimension. GNNs aim to learn node embeddings \( \mathbf{H}^{(l)} \) through the layer-specific transformations governed by parameters \( \theta^{(l)} \), expressed as \( \mathbf{H}^{(l)} = f_{\theta^{(l)}}(\mathbf{H}^{(l-1)}, \hat{\mathbf{A}}), l = 1,2,..., L \), where $L$ represents the number of layers, with the initial embedding \( \mathbf{H}^{(0)} = \mathbf{X} \).

\textbf{Sampling} is employed in GNNs to manage computational complexity, where a subset \( V^{'} \subset V \) of nodes is selected based on certain sampling rules \(\mathcal{P}\), such as importance sampling and Monte Carlo estimation~\citep{chen2018fastgcn}. Thus, The embeddings are estimated as \( f_{\theta^{(l)}}(\mathbf{H}_{[V^{'}]}^{(l-1)}, \hat{\mathbf{A}}) \), where \([V^{'}]\) denotes the indices corresponding to \(V^{'}\), reducing the data need to be processed. However, existing sampling algorithms have increasingly complicated the computation of \(\mathcal{P}\) to achieve more accurate approximations, leading to a growing overhead in sampling time.

\textbf{Compressed Sensing (CS)} is originally developed for sampling and reconstructing signal~\citep{candes2006near}. This technique can be applied to $\mathbf{H}^{(l)}$ if it exhibits sparsity in some specific transform domain. If $\mathbf{H}^{(l)} = \mathbf{U}^{(l)}\hat{\mathbf{H}}^{(l)}$, where $\hat{\mathbf{H}}^{(l)}$ is sparse, i.e., $\hat{\mathbf{H}}$ contains at most \( k \) non-zero rows, noted as $\Vert \hat{\mathbf{H}}^{(l)} \Vert_{0, row} \leq k$. The indices corresponding to the dense rows in \( \hat{\mathbf{H}}^{(l)} \) is called support, which holds the most significant information and can be used to effectively reconstruct \( \mathbf{H}^{(l)} \). The orthonormal basis \( \mathbf{U}^{(l)}\in \mathbb{R}^{N \times N} \) is used to transform \( \mathbf{H}^{(l)} \) into the sparse domain: \( \hat{\mathbf{H}}^{(l)} = \mathbf{U}^{T}\mathbf{H}^{(l)} \). Its existence is a necessary condition of CS. Fortunately, the orthonormal basis \( \mathbf{U}^{(l)} \) that satisfying \( \mathbf{H}^{(l)} = \mathbf{U}^{(l)}\hat{\mathbf{H}}^{(l)} \) where \( \Vert \hat{\mathbf{H}}^{(l)} \Vert_{0, row} \leq k \), always exists~\citep{isufi2024graph, bo2023survey} and can be derived from the graph’s structural properties~\citep{tsitsvero2016signals, puy2018random, chen2015discrete}, i.e., normalized Laplacian matrix. Let \( \mathbf{T}^{(l)} \in \mathbb{R}^{M \times d} \) where \( M \ll N \), be the measurement matrix, computed as:
\begin{equation}\label{eq:Tcomputation}
    \mathbf{T}^{(l)} = \mathbf{\Phi}^{(l)} \mathbf{U}^{(l)} \hat{\mathbf{H}}^{(l)}
\end{equation}
Here, \(\mathbf{\Phi}^{(l)}\in \mathbb{R}^{M \times N} \) is known as the sampling matrix. The measurement matrix represents the specific numerical values that can be directly observed during the computational process. To reconstruct the original sparse \( \hat{\mathbf{H}}^{(l)} \), the following optimization problem need to be solved: 
\begin{equation}\label{eq: optimization}
    \tilde{\mathbf{H}}^{(l)} = \text{min}_{\hat{\mathbf{H}}^{(l)}} \Vert \hat{\mathbf{H}}^{(l)} \Vert_{2,1} \quad \text{subject to}\quad \mathbf{T}^{(l)} = \mathbf{\Phi}^{(l)} \mathbf{U}^{(l)} \hat{\mathbf{H}}^{(l)}
\end{equation}
where $\Vert\cdot \Vert_{2,1}$ is $l_{2,1}$ norm~\citep{liu2018l_}. Accurate reconstruction requires that the matrix \( \mathbf{\Phi}^{(l)} \mathbf{U}^{(l)} \) satisfies the Restricted Isometry Property (RIP)~\citep{candes2005decoding}, formulated as 
\begin{equation}\label{eq:original_RIP}
    (1-\delta_k)\Vert \hat{\mathbf{H}}^{(l)} \Vert^2_F \leq \Vert {\mathbf{\Phi}^{(l)} \mathbf{U}^{(l)} \hat{\mathbf{H}}^{(l)}} \Vert^2_F \leq  (1+\delta_k)\Vert \hat{\mathbf{H}}^{(l)} \Vert^2_F
\end{equation}
where \( 0 < \delta_k < 1 \), and $\Vert \cdot \Vert_F$ is the Frobenius norm. After obtaining \( \tilde{\mathbf{H}}^{(l)} \) through Equation~(\ref{eq: optimization}), the original $\mathbf{H}^{(l)}$ can be reconstructed as:
\begin{equation}\label{eq: recon}
\mathbf{H}^{(l)} = [\mathbf{U}^{(l)}]^{T} \tilde{\mathbf{H}}^{(l)}
\end{equation}

\section{Compressed Sensing as Sampling in GNNs}
As discussed in Section~\ref{sec: pre}, CS reduces the amount of data required for computation by transforming \(\mathbf{H}^{(l)} \in \mathbb{R}^{N \times d}\) into a much smaller matrix \(\mathbf{T} \in \mathbb{R}^{M \times d}\) since \(M \ll N\). This reduction depends on converting \(\mathbf{H}^{(l)}\) into a sparse domain where its basis is $\mathbf{U}$, resulting in \(\hat{\mathbf{H}}^{(l)}\), such that \(\Vert \hat{\mathbf{H}}^{(l)} \Vert_{0, row} \leq k\). Efficient reconstruction is possible if such orthonormal basis \(\mathbf{U}\) and sampling matrix \(\mathbf{\Phi}\) exist and satisfy the RIP (Equation~(\ref{eq:original_RIP})).

When applied to GNNs, CS offers two main advantages over other schemes: (1) \(\mathbf{H} \in \mathbb{R}^{N \times d}\) can be sampled into a much smaller \(\mathbf{T} \in \mathbb{R}^{M \times d}\), significantly reducing computation time while retaining essential information; (2) CS enables lossless reconstruction at the output layer, allowing \(\mathbf{T}\) to be accurately expanded back to \(\mathbf{H}\) with high probability, as if all nodes were involved in the computation. Thus, a smaller sampled set can emulate the full training set, achieving high accuracy and reduced sampling time. This lossless property ensures that the model retains all information, thereby enhancing accuracy. Specifically:
\begin{equation}\label{eq:original_CS}
\mathbf{H}^{(l)}=f_{\theta^{(l)}}\left( Rec\left\{ \mathbf{T}^{(l-1)}\right\}, \hat{\mathbf{A}} 
    \right)
\end{equation}
where $Rec\{\cdot\}$ represents the processing of reconstruction (Equation~(\ref{eq: optimization}) and Equation~(\ref{eq: recon})). However, the iterative processes in Equation~(\ref{eq:original_CS}) is highly inefficient and has the following challenges:
\begin{itemize}[topsep=0pt,itemsep=-1ex,partopsep=1ex,parsep=1ex, leftmargin=*]
    \item \textbf{Expensive Computations of $\mathbf{U}^{(l)}$ and $\mathbf{\Phi}^{(l)}$.} Determining appropriate orthonormal bases \( \mathbf{U}^{(l)} \) and sampling matrices \( \mathbf{\Phi}^{(l)} \) for \( l = 1, ..., L \), is time-consuming. While Section~\ref{sec: pre} theoretically confirms the existence of \( \mathbf{U}^{(l)} \), practical computation is costly since it requires matrix decompositions with an average time complexity of \( O(n^3) \). More seriously, \( \mathbf{H}^{(l)} \) changes across GNN layers, therefore a fixed \( \mathbf{U}^{(l)} \) is unlikely to meet the sparsity requirements for all layers, which necessitating \( (L + 1) \) separate decompositions. Similarly, \( \mathbf{\Phi}^{(l)} \) must adapt to changes in \( \mathbf{U}^{(l)} \) to maintain RIP, requiring an additional \( (L + 1) \) adjustments. In summary, determining \( \mathbf{U}^{(l)} \) and \( \mathbf{\Phi}^{(l)} \) involves \( 2(L + 1)O(n^3) \) costly computations during training.
    \item \textbf{Accurate but Time-inefficient Reconstruction.} As in Equation~(\ref{eq:original_CS}), to minimize error propagation, we reconstruct \( \mathbf{H} \) at every layer before proceeding to the next layer. However, this incurs significant computational overhead. The fastest known reconstruction algorithm has an average time complexity of \( O(nm) \)~\citep{maleki2010approximate}, where \( n \) is the signal dimension and \( m \) is the measurement length. For GNNs, this translates to an average reconstruction time complexity of \( O(dM) \) per layer, resulting in a total cost of \( O(dML) \) for an \( L \)-layer GNN. Such overhead greatly reduces training efficiency.
\end{itemize}

Consequently, directly applying CS to GNN sampling introduces significant time complexities. To effectively integrate CS into GNNs and ensure its efficiency, we must overcome the two obstacles:
\begin{enumerate}[topsep=0pt,itemsep=-1ex,partopsep=1ex,parsep=1ex, leftmargin=*]
    \item[I.] \textbf{Working with Unknown \( \mathbf{U}^{(l)} \) and Universal \( \mathbf{\Phi} \).} Given the high computational cost of determining \( \mathbf{U}^{(l)} \), we need to satisfy or approximate CS's necessary and sufficient condition without explicitly knowing \( \mathbf{U}^{(l)} \). Without \( \mathbf{U}^{(l)} \), identifying the support and determining essential nodes for reconstruction becomes challenging, complicating the construction of \( \mathbf{\Phi}^{(l)} \). Since \( \mathbf{\Phi}^{(l)} \) is layer-specific, calculating it for each layer is impractical. Thus, we require a method that works with an unknown \( \mathbf{U} \) using a universal sampling matrix \( \mathbf{\Phi} \), ensuring \( \mathbf{\Phi} \) remains adaptable to any \( \mathbf{U} \) while satisfying compressed sensing conditions.
    \item[II.] \textbf{Balancing computational efficiency with the need for accurate reconstruction.} If we sample once at the input layer and use these results throughout the GNN computation, followed by reconstruction only at the output layer, this approach requires just one sampling and reconstruction step for the entire training process. Although it may introduce some accuracy loss due to reduced intermediate layer information, it remains efficient if this loss is controllable with a known upper bound, allowing a balance between computational efficiency and model accuracy.
\end{enumerate}

\section{Methodology}\label{sec: method}
The overall YOSO algorithm is presented in Section~\ref{sec: ouralg}, where we also address the challenge of working with the unknown \( \mathbf{U} \). Followed by design of universal sampling matrix $\mathbf{\Phi}$ in Section~\ref{sec: transform}.

\subsection{YOSO Design}\label{sec: ouralg}
\begin{algorithm}[!b]
\small
\caption{Forward and Backward Propagation of YOSO}
\label{forward_backward}
\begin{algorithmic}[1]
\STATE Initialize $\Theta$, $\mathbf{U}$, and $\hat{\mathbf{H}}^{(L)}$
\WHILE{not converged}
    \STATE Compute $\mathbf{T}^{(0)} = \mathbf{\Phi} \mathbf{U} \mathbf{\hat{X}}$
    \FOR{$l = 1$ to $L - 1$}
        \STATE Compute $\mathbf{T}^{(l)} = \sigma \left( \mathbf{\Phi} \hat{\mathbf{A}} \mathbf{W}^{(l)} \mathbf{T}^{(l-1)} \right)$
    \ENDFOR
    \STATE Compute $\mathbf{Z} = \sigma \left( \mathbf{\Phi} \hat{\mathbf{A}} \mathbf{W}^{(L)} \mathbf{T}^{(L-1)} \right)$
    \STATE Compute reconstruction Loss: $\mathcal{L}_{\text{recon}} = \frac{1}{2} \Vert \mathbf{Z} - \mathbf{\Phi} \mathbf{U} \hat{\mathbf{H}}^{(L)} \Vert_F^2 + \lambda \Vert \hat{\mathbf{H}}^{(L)} \Vert_{2,1}$
    \STATE Compute GNN Loss: $\mathcal{L}_{GNN}^{\Theta}(\mathbf{Z})$
    \STATE Compute Total Loss: $\mathcal{L} = \alpha \mathcal{L}_{\text{recon}} + \beta \mathcal{L}_{GNN}^{\Theta}(\mathbf{Z})$
    \STATE Compute gradient w.r.t $\Theta$: $\nabla_{\Theta} \mathcal{L} = \alpha \nabla_{\Theta} \mathcal{L}_{\text{recon}} + \beta \nabla_{\Theta} \mathcal{L}_{GNN}^{\Theta}(\mathbf{Z})$
    \STATE Compute gradient w.r.t $\mathbf{U}$: $\nabla_\mathbf{U} \mathcal{L} = \alpha \nabla_\mathbf{U} \mathcal{L}_{\text{recon}} + \beta \nabla_\mathbf{U} \mathcal{L}_{GNN}^{\Theta}(\mathbf{Z})$
    \STATE Compute gradient w.r.t $\hat{\mathbf{H}}^{(L)}$: $\nabla_{\hat{\mathbf{H}}^{(L)}} \mathcal{L} = \eta_{\hat{\mathbf{H}}^{(L)}} \nabla_{\hat{\mathbf{H}}^{(L)}} \mathcal{L}_{\text{recon}}$
    \STATE Update $\Theta$: $\Theta \leftarrow \Theta - \eta_{\Theta} \nabla_{\Theta} \mathcal{L}$
    \STATE Update $\mathbf{U}$: $\mathbf{U}_{\text{temp}} = \mathbf{U} - \eta_\mathbf{U} \nabla_\mathbf{U} \mathcal{L}$
    \STATE Project $\mathbf{U}$ onto the Stiefel manifold~\citep{koochakzadeh2016nonnegative} to ensure $\mathbf{U}^T\mathbf{U} = I$
    \STATE Update $\hat{\mathbf{H}}^{(L)}$: $\hat{\mathbf{H}}^{(L)} \leftarrow \hat{\mathbf{H}}^{(L)} - \eta_{\hat{\mathbf{H}}} \nabla_{\hat{\mathbf{H}}^{(L)}} \mathcal{L}$
\ENDWHILE
\end{algorithmic}
\end{algorithm}
YOSO proposes a CS-based sampling and reconstruction framework, where nodes are sampled once at the input layer, followed by a lossless reconstruction at the output layer during each epoch. As shown in Algorithm~\ref{forward_backward}, the entire training process of YOSO consists of forward propagation, loss computation, and backward propagation, similar to the conventional GNN training. Unlike the standard process, YOSO operates within a specific sparse domain instead of the original data domain. Initially, the data is transformed into the sparse domain (Line 3), where the one-time sampling is also performed by using the sampling matrix \( \mathbf{\Phi} \). The subsequent steps--forward propagation (Lines 4-7), loss computation (Lines 8-10), and backward propagation (Lines 11-17)--are all executed within this sparse domain. The detailed description is as follows:\\
\textbf{One time sampling (Line 3).} Given a graph $G=\{ V,E,\hat{\mathbf{A}}, \mathbf{X} \}$, where specific $\hat{\mathbf{A}}$ is the normalized Laplacian matrix. We perform the sampling stage only once using the sampling matrix \( \mathbf{\Phi} \) on the sparsity domain \(\hat{\mathbf{X}}\) as $\mathbf{\Phi}\mathbf{U}\hat{\mathbf{X}}$, resulting in \( \mathbf{T}^{(0)} \in \mathbb{R}^{M \times d} \), where \( M \ll |V| = N \). This process involves the construction of the sampling matrix \( \mathbf{\Phi} \), for details, please refer to Section~\ref{sec: transform}.\\
\textbf{Forward propagation (Lines 4-7).} The forward propagation of YOSO can be expressed as:
\begin{equation}\label{eq: forward}
\left\{
\begin{array}{lr}
\mathbf{T}^{(l)} = \sigma \left( 
\mathbf{\Phi}\hat{\mathbf{A}}\mathbf{W}^{(l)}\mathbf{T}^{(l-1)}
\right)  & 1 \leq l\leq L-1 \\
\mathbf{U}, \hat{\mathbf{H}}^{(L)} = Rec\left\{\sigma\left(
\mathbf{\Phi}\hat{\mathbf{A}}\mathbf{W}^{(L)}\mathbf{T}^{(L-1)}\right)\right\} & l=L
\end{array}
\right.
\end{equation}
where $\sigma(\cdot)$ is the activation function, $\mathbf{W}^{(l)}, l=1,..., L$ is the $l-$th layer's trainable parameters, $\mathbf{U}$ is the unknown orthonormal basis and the method for addressing this (working with unknown $\mathbf{U}$) will be discussed in the following.\\
\textbf{Loss function and working with unknown $\mathbf{U}$ (Lines 8-10).} First, we discuss the construction of YOSO's loss function in (1), and then in (2), we explain why this construction effectively addresses the challenge of working with the unknown \( \mathbf{U} \).\\
\textbf{(1) Loss function.} The $Rec\{ \cdot\}$ in Equation~(\ref{eq: forward}) is equal to solve the following optimization problem:
\begin{equation}\label{eq:revised_opt}
\underset{\hat{\mathbf{H}}^{(L)}, \mathbf{U}}{\min} \ \frac{1}{2} \left\Vert \mathbf{Z} - \mathbf{\Phi} \mathbf{U} \hat{\mathbf{H}}^{(L)} \right\Vert_{F}^{2} + \lambda \left\Vert \hat{\mathbf{H}}^{(L)} \right\Vert_{2,1} \quad \text{s.t. } \mathbf{U}\mathbf{U}^{T} = \mathbf{U}^{T}\mathbf{U} = \mathbf{I}
\end{equation}
where \( \mathbf{Z} = \sigma(\mathbf{\Phi}\hat{\mathbf{A}}\mathbf{W}^{(L)}\mathbf{T}^{(L-1)}) \) represents the sampled measurement matrix at output layer, and \( \lambda \) is a hyperparameter controlling the balance between data fidelity and sparsity. Equation~(\ref{eq:revised_opt}) is a non-trivial optimization problem involving both \( \hat{\mathbf{H}}^{(L)} \) and \( \mathbf{U} \)  due to non-convexity introduced by orthogonality constraint ($\mathbf{U}\mathbf{U}^{T} = \mathbf{U}^{T}\mathbf{U} = \mathbf{I}$) and the interaction between variables. To overcome it, we perform joint optimization of Equation~(\ref{eq:revised_opt}) with the GNN's specific loss function (e.g., cross-entropy). Let the GNN's loss function be \( \mathcal{L}_{GNN}^{\Theta}\), where \( \Theta = \{\mathbf{W}^{(1)},..., \mathbf{W}^{(L)} \} \) represents the set of all trainable parameters. The joint optimization objective function is defined as:
\begin{equation}\label{eq:total_loss}
     \underset{\hat{\mathbf{H}}^{(L)}, \mathbf{U}, \Theta}{\min} \left\{   \alpha \left(
    \frac{1}{2} \left\Vert \mathbf{Z} - \mathbf{\Phi} \mathbf{U} \hat{\mathbf{H}}^{(L)} \right\Vert_{F}^{2} + \lambda \left\Vert \hat{\mathbf{H}}^{(L)} \right\Vert_{2,1}
    \right) + \beta \mathcal{L}_{GNN}^{\Theta}(\mathbf{Z}) \right\} \quad \text{s.t. } \mathbf{U}\mathbf{U}^{T} = \mathbf{U}^{T}\mathbf{U} = \mathbf{I}
\end{equation}
where $\alpha$ and $\beta$ is the hyperparameters to balance the reconstruction loss and GNN loss.\\
\textbf{(2) Working with unknown} $\mathbf{U}$. To address the challenge of unknown \( \mathbf{U} \), we treat \( \mathbf{U} \) as an optimization target. Using Equation~(\ref{eq:total_loss}), we obtain a total loss, which is then used to generate gradients for updating $\mathbf{U}$ through all training process (Detailed calculation of the gradient of the loss in Equation~(\ref{eq:total_loss}) with respect to \( \mathbf{U} \) can be found in Appendix~\ref{app:gradient}).\\
\textbf{Backward Propagation (Lines 11-17).} The backward propagation process uses the loss generated by Equation~(\ref{eq:total_loss}) to update three parameters, which are $\mathbf{U}$, $\hat{\mathbf{H}}^{(L)}$, and $\Theta=\{ \mathbf{W}^{(1)}, \dots, \mathbf{W}^{(L)} \}$ through gradient descent. This process results in three gradients, namely \( \nabla_\mathbf{U} \mathcal{L} \), \( \nabla_{\hat{\mathbf{H}}^{(L)}} \mathcal{L} \), and \( \nabla_{\Theta} \mathcal{L} \), each corresponding to three learning rates \( \eta_{\mathbf{U}} \), \( \eta_{\hat{\mathbf{H}}^{(L)}} \), and \( \eta_{\Theta} \), respectively. For the detailed setting of hyperparameters used here, i.e., $\alpha$ and $\eta_{\mathbf{U}}$, please refer to Appendix~\ref{app: hyperparameter} and the detailed gradient computation list in Appendix~\ref{app:gradient}.

Through Algorithm~\ref{forward_backward}, we obtain both \( \mathbf{U} \) and \( \hat{\mathbf{H}}^{(L)} \). With \( \mathbf{U} \) now determined, we can apply Equation~(\ref{eq: recon}) to reconstruct \( \mathbf{H}^{(L)} \), which can then be utilized for downstream tasks, such as link prediction. Compared to Equation~(\ref{eq:original_CS}), the process described in Algorithm~\ref{forward_backward} trades some accuracy for improved efficiency, and importantly, this accuracy loss is bounded. For detailed statements and proofs, please refer to Appendix~\ref{app:error_bound}.

\subsection{Construction of Sampling Matrix \( \mathbf{\Phi} \)}\label{sec: transform}
When the orthonormal basis \( \mathbf{U} \) remains unspecified before training, we encounter the challenge of computing \( \mathbf{T}^{(0)} = \mathbf{\Phi} \mathbf{U} \mathbf{\hat{X}} \) in Equation~(\ref{eq: forward}) since the absence of knowledge about \( \mathbf{U} \) complicates the design of \( \mathbf{\Phi} \). In traditional CS, \( \mathbf{U} \) maps data into a sparse domain where the support (i.e., the indices of non-zero rows) is clearly identifiable, and these non-zero rows contain the crucial information. This clarity allows \( \mathbf{\Phi} \) to be designed in a targeted manner based on the support. Without knowing \( \mathbf{U} \), it becomes challenging to design a \( \mathbf{\Phi} \) that effectively captures the essential information. Therefore, the main difficulty lies in designing an effective and universal sampling matrix \( \mathbf{\Phi} \) that not only accurately captures the essential characteristics of the graph data but also works with any \( \mathbf{U} \) without violating the RIP. 

To address challenge, we propose an approach that integrates the design of a matrix \( \hat{\mathbf{S}} \in \mathbb{R}^{M \times N} \), derived from the graph structure, with the construction of the sampling matrix \( \mathbf{\Phi} \), i.e., \( \mathbf{\Phi} = \hat{\mathbf{S}}\otimes \mathbf{\Sigma} \) where $\mathbf{\Sigma}\in \mathbb{R}^{M \times N}$ is a random matrix and $\otimes$ is element-wise production. 

$\hat{\mathbf{S}}$ remains unchanged through the entire training and is determined only once during pre-processing phase. Design of $\hat{\mathbf{S}}$ is graph-structure-based for two reasons: first, the graph structure is invariant, and second, it reflects the importance of certain nodes, which is crucial for the GNN message-passing process. For the sampling matrix \( \mathbf{\Phi} \), it is essential to be row full rank. Intuitively, \( \mathbf{\Phi} \) serves to linearly combine the features or embeddings of nodes according to weights corresponding to the indices of the support (non-zero rows). If \( \mathbf{\Phi} \) is row over-ranked, it results in redundant information, whereas a row under-ranked \( \mathbf{\Phi} \) leads to information loss. Thus, ensuring a row full-rank sampling matrix is crucial for effectively capturing the necessary information.\\
\textbf{Construction of $\hat{\mathbf{S}}$. } Considering the normalized Laplacian matrix \( \hat{\mathbf{A}} = \mathbf{I} - \mathbf{D}^{-1/2} \mathbf{A} \mathbf{D}^{-1/2} \) where $\mathbf{D}$ and $\mathbf{A}$ are the degree matrix and adjacency matrix, respectively. The \( N \) nodes correspond to \( N \) eigenvalues from $\hat{\mathbf{A}}$'s spectral decomposition, denoted as \( \{\lambda_1, \ldots, \lambda_N\} \) and $\lambda_i \geq 0$ for any $i$ holds. These eigenvalues often reflect the important structural properties of the graph. For example, larger eigenvalues correspond to more influential nodes within the graph. To construct the sampling probability distribution, we define \( P(i) = \frac{\lambda_i}{\sum_{j=1}^N \lambda_j} \), where node \( i \) has a sampling probability proportional to its eigenvalue relative to the total eigenvalue sum. Using this probability distribution, we sample \( M \) times to form the \( M \) rows of \( \hat{\mathbf{S}} \). Suppose node \( i \) is sampled; the corresponding row in \( \hat{\mathbf{S}} \) will include node \( i \)'s 1-hop neighbors. Assume node \( i \) has \( N(i) \) neighbors, each neighbor is randomly sampled with a probability of \( \frac{1}{N(i)} \). This construction ensures that \( \hat{\mathbf{S}} \) will not contain any all-zero rows, thanks to the self-loop added by the normalized Laplacian. Consequently, the matrix \( \mathbf{\Phi} = \hat{\mathbf{S}}\mathbf{\Sigma} \) will be row full rank (detailed proof in Appendix~\ref{app:full_rank}), avoiding any issues with row rank deficiency.\\
\textbf{Construction of $\mathbf{\Sigma}$.} Some studies have highlighted the importance of randomness in achieving the RIP~\citep{baraniuk2008simple}. Therefore, we define \( \Sigma \) as a random matrix. Intuitively, since we do not have precise knowledge of the support, we randomly sample \( M \) nodes based on eigenvalue weights to estimate the support. The matrix \( \mathbf{\Sigma} \) should reflect the contribution level of each node \( i \) to the non-zero rows (i.e., the support). For instance, if node \( k \) is shared by both nodes \( i \) and \( j \), we need to determine how much node \( k \) contributes to node \( i \) and to node \( j \). This is crucial for ensuring accurate reconstruction and satisfying the RIP. For any column \( j \) in \( \hat{\mathbf{S}} \), assume it contains \( g(j) \) non-zero elements. We assign the corresponding elements in \( \Sigma \) as random values drawn from a Gaussian distribution \( N(0, \frac{1}{g(j)}) \). This design helps capture the contribution levels effectively, which is important for achieving the Restricted Isometry Property (detailed proof in Appendix~\ref{app:Phi_RIP}). 

\section{Related Work}
A widely accepted criterion~\citep{liu2021sampling} divides current different sampling methods into three categories: node-wise sampling, layer-wise sampling, and subgraph-based sampling, depending on the granularity of the sampling operation during mini-batch generation. \newline
\textbf{Node-wise Sampling:} This fundamental approach, pioneered by works such as GraphSage~\citep{hamilton2017inductive} and others~\citep{ying2018graph, chen2017stochastic, dai2018learning}, involves sampling at the individual node level. Each node's neighbors are selected according to specific probabilities, often using a uniform distribution. For example, GraphSage samples $k-$hop neighbors at varying depths, with the sampling size, or fanout, for each depth tailored to optimize model performance. This approach, while simple and effective, has been criticized for its exponential increase in sampling time complexity as the number of GNN layers grows.\newline
\textbf{Layer-wise Sampling:} Developed to address the exponential growth in computational complexity as network depth increases in node-wise sampling, this method samples multiple nodes simultaneously in one step. Techniques like FastGCN~\citep{chen2018fastgcn} reframe GNN loss functions as integral transformations and utilize importance sampling and Monte-Carlo approximation to manage variance. Further developments, such as AS-GCN~\citep{huang2018adaptive} and LADIES~\citep{zou2019layer}, focus on maintaining sparse connections between sampled nodes to aid convergence. However, these methods tend to introduce additional complexity and computational cost.\newline
\textbf{Subgraph-based Sampling:} These methods form mini-batch training subgraphs using graph partitioning algorithms. Cluster-GCN~\citep{chiang2019cluster} partitions the full graph into clusters, sampling these clusters to create subgraphs for training batches. GraphSAINT~\citep{zeng2019graphsaint} dynamically estimates sampling probabilities for nodes and edges to form subgraphs over which the full GNN model is trained. While these techniques typically improve model accuracy, they also lead to longer training time.

\section{Experiments}
In Section~\ref{sec: overall}, we evaluate the training time along with model accuracy across two learning tasks: node classification and link prediction. Also, to investigate convergence performance, we assess the convergence of both the baselines and YOSO in Section~\ref{sec: convergence}. Finally, we conduct an ablation study on the proposed compensations in Section~\ref{sec: ablation}. Details on the dataset, baselines, experimental hardware and software configuration can be found in Section~\ref{sec: experimental setting} and Appendix~\ref{app: hard/software}.
\subsection{Experimental Settings}\label{sec: experimental setting}
\textbf{Datasets. }For the node classification task, we selected Reddit~\citep{hamilton2017inductive}, ogbn-arxiv and ogbn-products~\citep{hu2020open}. For the link prediction task, we used ogbl-ppa, and ogbl-citation2~\citep{hu2020open}. For detailed dataset statistics, data splits and metrics, please refer to Appendix~\ref{app: datasets}. \newline
\textbf{Baselines and Implementation. } The baselines used in this paper include node-wise sampling methods (GraphSage~\citep{hamilton2017inductive} and VR-GCN~\citep{chen2017stochastic}), layer-wise sampling methods (FastGCN~\citep{chen2018fastgcn}, AS-GCN~\citep{huang2018adaptive} and LADIES~\citep{zou2019layer}) and subgraph-based sampling methods (Cluster-GCN~\citep{chiang2019cluster} and GraphSAINT~\citep{zeng2019graphsaint}). Notably, several baseline models lacked implementations for link prediction, prompting us to modify them accordingly. Detailed information on the source code for these baselines, the YOSO implementation, and other related materials can be found in Appendix~\ref{app: baselines}.\\
\textbf{Hyperparameter Setting.} All experiments are conducted using a two-layer GNN. Detailed hyperparameter settings are described in Appendix~\ref{app: hyperparameter}.
\subsection{Overall Comparison}\label{sec: overall}
In this section, we evaluate baselines and YOSO with two key metrics: model accuracy (varies with different datasets and tasks) and total training time. The training time is broken down into three non-overlapped parts: Sampling, Mem2GPU, and Computation.

\begin{figure}[!t]
\centering
    \begin{minipage}[b]{0.32\linewidth}
        \centering
        \includegraphics[width=\linewidth]{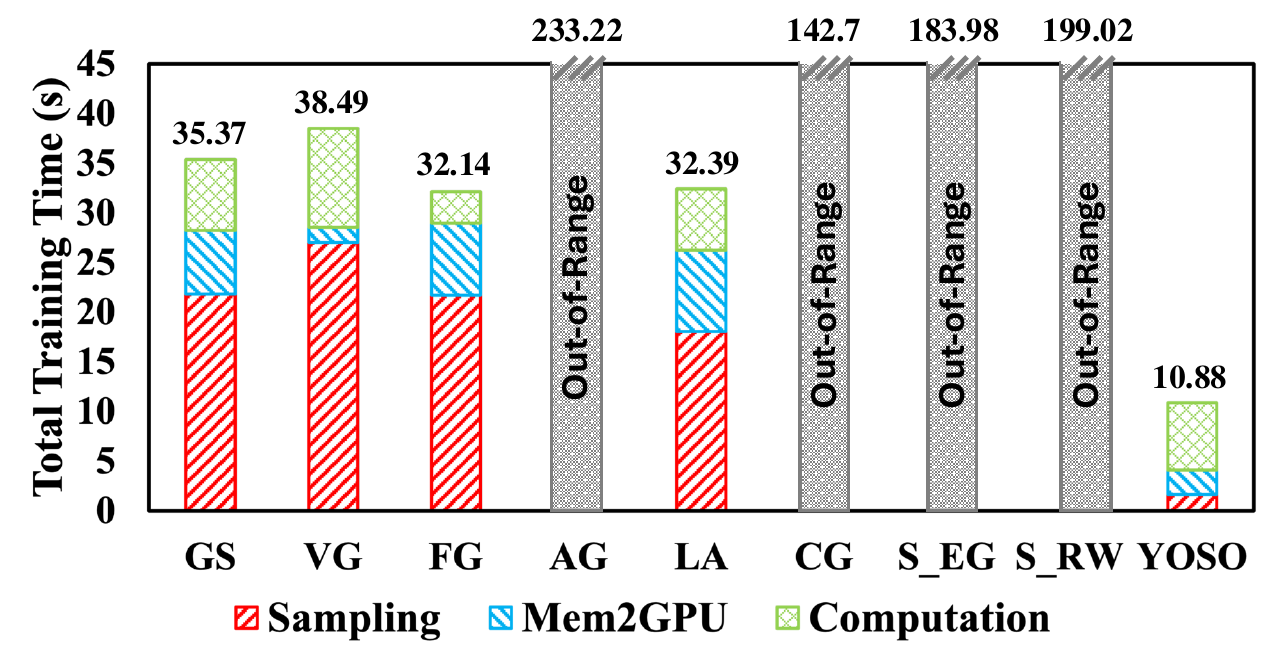}
        \subcaption{ogbn-arxiv}
    \end{minipage}
    \hfill
    \begin{minipage}[b]{0.32\linewidth}
        \centering
        \includegraphics[width=\linewidth]{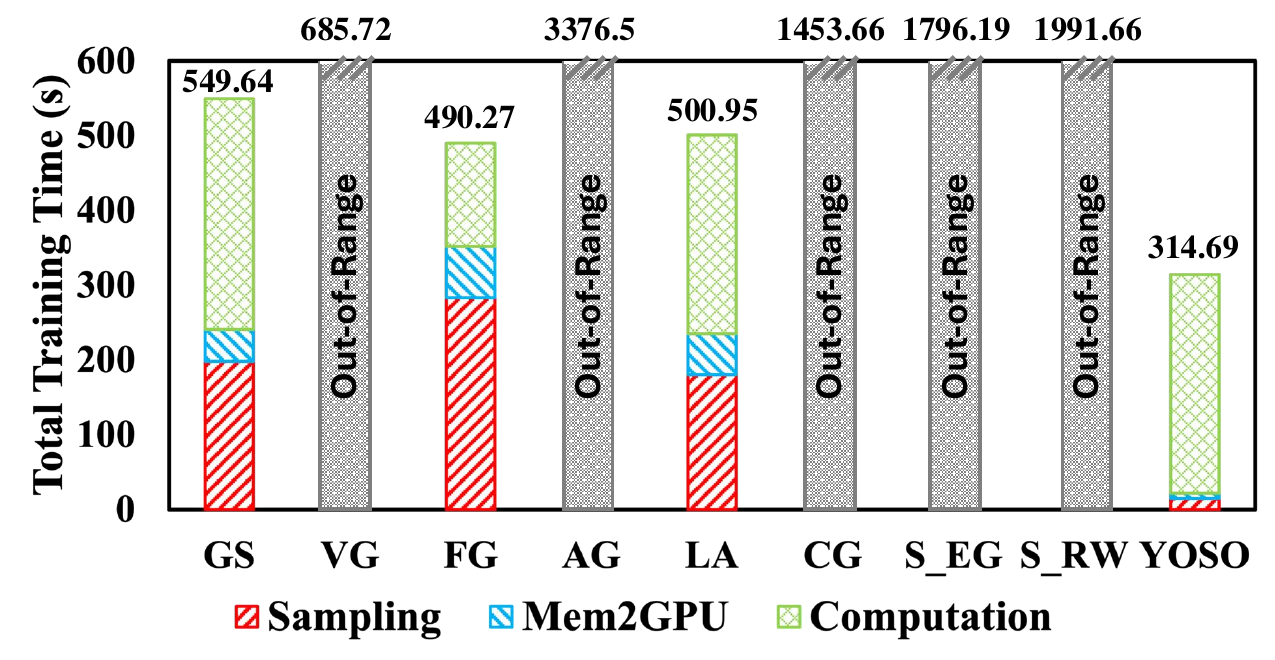}
        \subcaption{Reddit}
    \end{minipage}
    \hfill
    \begin{minipage}[b]{0.32\linewidth}
        \centering
        \includegraphics[width=\linewidth]{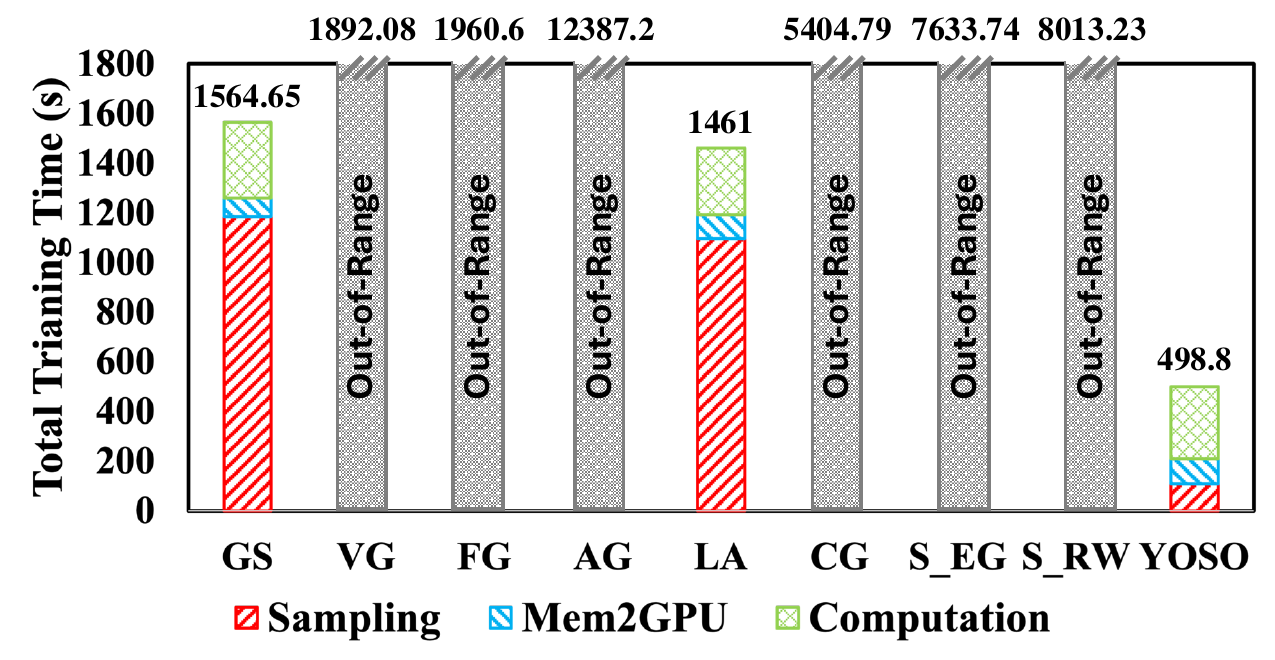}
        \subcaption{ogbn-products}
    \end{minipage}
    
    \vspace{5pt}

    \begin{minipage}[r]{0.48\linewidth}
        \begin{flushright}
        \includegraphics[width=0.667\linewidth]{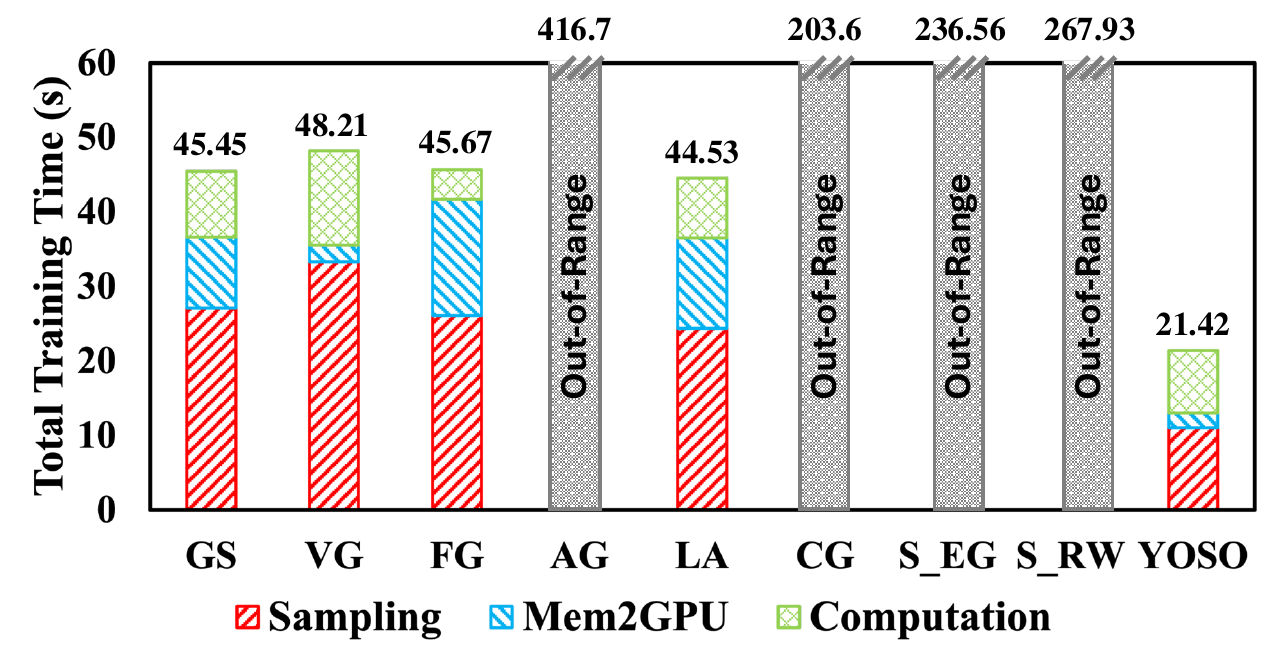}
        \subcaption{ogbl-ppa}
        \end{flushright}
    \end{minipage}
    \hfill
    \begin{minipage}[r]{0.48\linewidth}
        \begin{flushleft}   \includegraphics[width=0.667\linewidth]{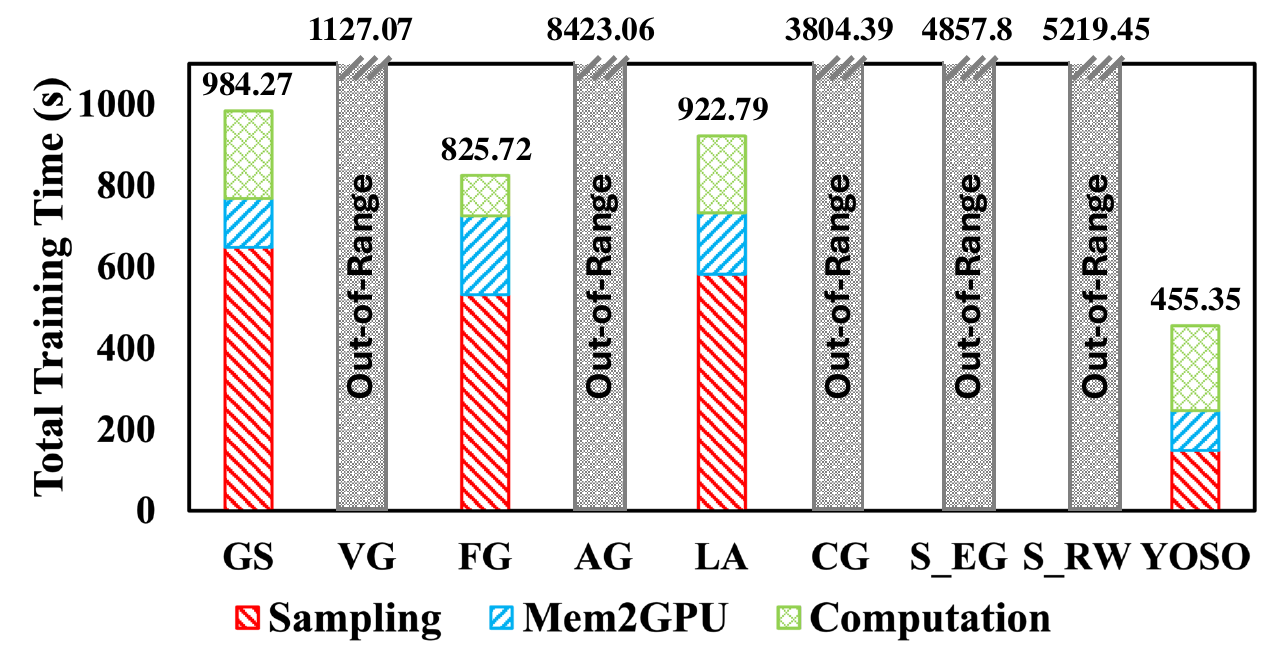}
        \subcaption{ogbl-citation2}
        \end{flushleft}
    \end{minipage}
\caption{Total training time comparison with the breakdown times including Sampling, Computation, and Mem2GPU. This evaluation covers two learning tasks across five datasets: (a) to (c) represent the results for the node classification task on ogbn-arxiv~\citep{hu2020open}, Reddit~\citep{hamilton2017inductive}, and ogbn-products~\citep{hu2020open}, respectively; while (d)-(e) correspond to the link prediction task on ogbl-ppa~\citep{hu2020open} and ogbl-citation2~\citep{hu2020open}. We use the same model name abbreviations as in Figure~\ref{fig:result_motivation}.}
\label{fig: all_time}
\end{figure}

\textbf{Node Classification Task:}
First, YOSO achieves the shortest total training time with an average of 75.3\% reduction across all datasets compared to all baselines as shown in Figure~\ref{fig: all_time}. For example, YOSO reduces around 95\% total training time from 233.22 seconds (ogbn-arxiv/AS-GCN) and 12,387.2 seconds (ogbn-products/AS-GCN) to 199.02 and 8,013.23 seconds, respectively. The main reason is that YOSO significantly reduces the sampling time while introducing a little re-construction overhead. As shown in Figure~\ref{fig: all_time}(a)-(c), the most substantial sampling time reduction occurs on the Reddit dataset, where YOSO achieved a 99\% decrease, cutting the sampling time from 1149.02 seconds for GraphSAINT-EDGE and 1107.54 seconds for Random Walk to just 15.13 seconds. 
On average, YOSO reduced sampling time by approximately 95.7\% compared to all other baselines. 

For model accuracy shown in Table~\ref{tab:model_acc}, YOSO consistently matches or closely approaches the top performers. For example, YOSO obtains an accuracy of 0.71 on ogbn-arxiv, just 0.01 below GraphSage. On Reddit, it achieves the highest score of 0.967, matching GraphSAINT-Random Walk, and on ogbn-products, it reaches 0.787, slightly trailing GraphSAINT-EDGE's 0.792. 

\begin{table}[t!]
\centering
\small
\caption{Model accuracy results for different sampling schemes on node classification and link prediction tasks. For specific evaluation metrics on each dataset, please refer to Table~\ref{tab:dataset}.}
\label{tab:model_acc}
\begin{tabular}{|c|ccccc|}
\hline
\multirow{3}{*}{\begin{tabular}[c]{@{}c@{}}Different\\ Sampling\\ Schemes\end{tabular}} &
  \multicolumn{5}{c|}{Dataset} \\ \cline{2-6} 
            & \multicolumn{3}{c|}{Node Classification}                            & \multicolumn{2}{c|}{Link Prediction} \\ \cline{2-6} 
            & ogbn-arxiv    & Reddit         & \multicolumn{1}{c|}{ogbn-products} & ogbl-ppa       & ogbl-citation2      \\ \hline
GraphSage   & \textbf{0.72}          & 0.949          & \multicolumn{1}{c|}{0.772}         & 0.1704         & \textbf{0.8054}              \\
VR-GCN      & 0.697         & 0.962          & \multicolumn{1}{c|}{0.699}         & 0.1704         & 0.7967              \\
FastGCN     & 0.438         & 0.927          & \multicolumn{1}{c|}{0.404}         & 0.1088         & 0.6555              \\
AS-GCN      & 0.687         & 0.964          & \multicolumn{1}{c|}{0.51}          & 0.1245         & 0.6593              \\
LADIES      & 0.649         & 0.927          & \multicolumn{1}{c|}{0.501}         & 0.1131         & 0.6693              \\
Cluster-GCN & 0.653         & 0.966          & \multicolumn{1}{c|}{0.769}         & 0.2053         & 0.7904              \\
GraphSAINT-EG &
  0.702 &
  \textbf{0.967} &
  \multicolumn{1}{c|}{\textbf{0.792}} &
  0.2143 &
  0.8039 \\
GraphSAINT-RW &
  0.701 &
  \textbf{0.967} &
  \multicolumn{1}{c|}{0.783} &
  \textbf{0.2263} &
  \textbf{0.8054} \\
YOSO        & \textbf{0.72} & \textbf{0.967} & \multicolumn{1}{c|}{0.787}         & 0.2238         & 0.8025              \\ \hline
\end{tabular}
\end{table}

\textbf{Link Prediction Task:}
For total training time, similar to the node classification task, YOSO achieves the best training time with a 72.13\% average training time decrease across all datasets for the link prediction. For example, YOSO  decreases the training time for the ogbl-ppa dataset from 44.53 seconds with AG-GCN to 21.42 seconds, and for the ogbl-citation2 dataset from 8423.06 seconds with AG-GCN to 455.35 seconds.This improvement is consistent with the node classification task, where YOSO achieves considerable reductions in sampling time while introducing minimal reconstruction overhead. As depicted in Figure~\ref{fig: all_time}(d)-(e), YOSO achieves an average sampling time reduction of about 80.5\% across all datasets. As for model accuracy, outlined in Table~\ref{tab:model_acc}, YOSO maintained results with only a very small gap--0.0025 on ogbn-arxiv and 0.0029 on ogbl-citation2--compared to the best results achieved by GraphSAINT-Random Walk and GraphSage, respectively.

In summary, for both tasks of node classification and link prediction, by combining high accuracy with substantial reductions in sampling and total training time, YOSO demonstrates its efficiency in GNN training and significantly improves both sampling and total training times across all datasets while maintaining competitive accuracy, highlighting its effectiveness compared to the baselines on the node classification task.

\subsection{Convergence Comparison}\label{sec: convergence}
\begin{figure}[!t]
\begin{minipage}{0.5\linewidth}
    \centerline{\includegraphics[width=\linewidth]{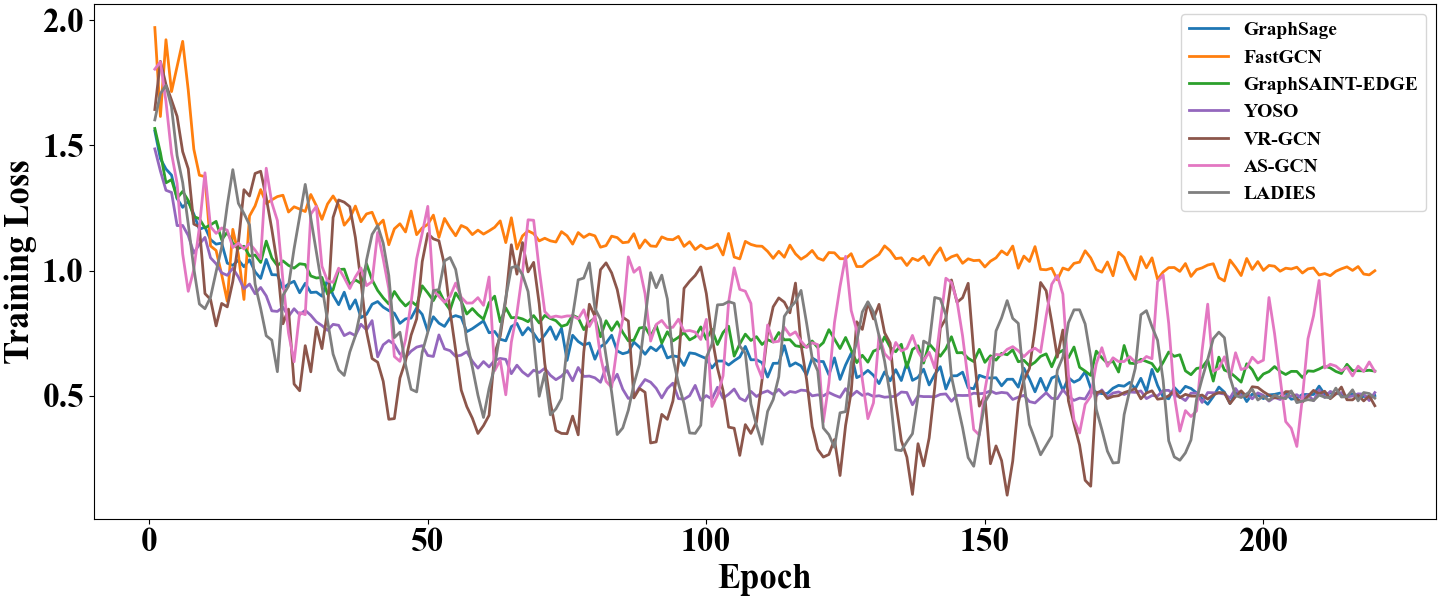}}
    \subcaption{ogbn-arxiv}
\end{minipage}
\begin{minipage}{0.5\linewidth}
    \centerline{\includegraphics[width=\linewidth]{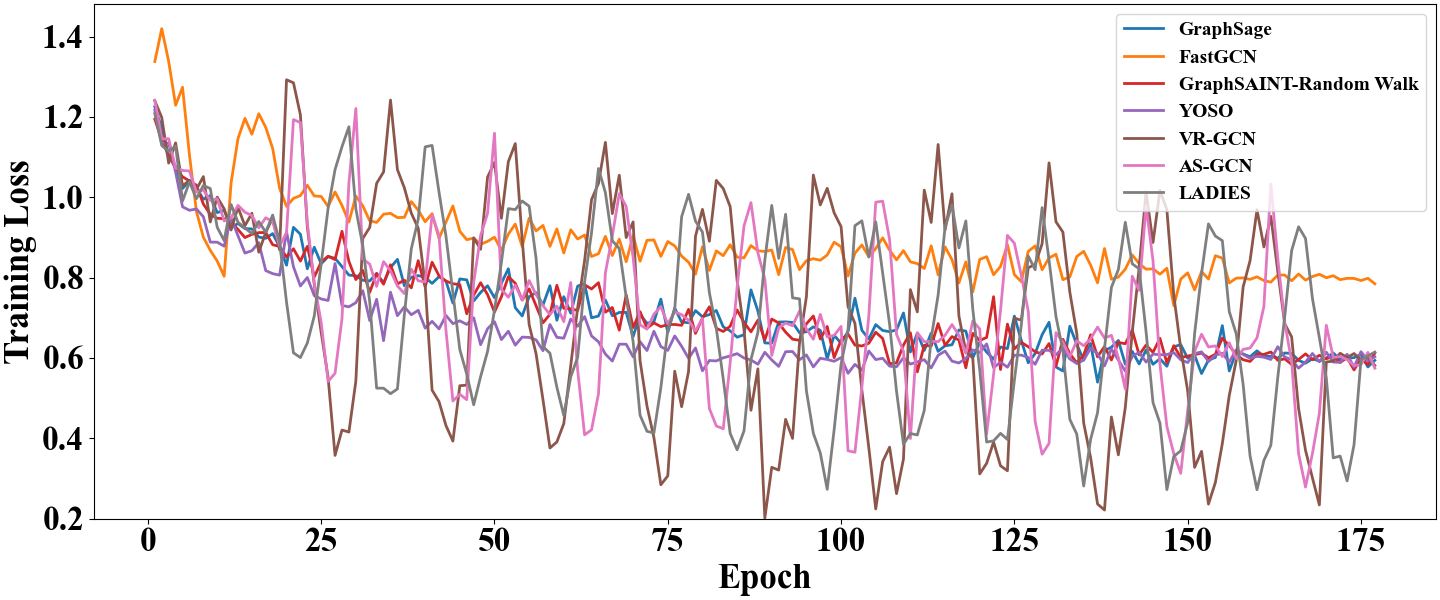}}
    \subcaption{ogbl-ppa}
\end{minipage}
\vspace{-2mm}
\caption{Training loss and epoch curves for YOSO and baselines on two benchmark datasets.}
\label{fig:convergence}
\end{figure}

We investigate YOSO's convergence performance compared to other baselines. Specifically, we select ogbn-arxiv and ogbl-ppa as representatives for node classification and link prediction, respectively. The training loss-epoch curves are shown in Figure~\ref{fig:convergence}.

In both experiments, YOSO consistently outperformed the baselines in terms of convergence speed and stability. On the ogbn-arxiv dataset, YOSO reached a lower training loss more rapidly than GraphSAGE, GraphSAINT-EDGE, and FastGCN, with significantly fewer oscillations, indicating a more stable and efficient training process. Similarly, on the ogbl-ppa dataset, YOSO demonstrated faster convergence and maintained a smoother training loss curve, while the baselines, especially FastGCN, exhibited more fluctuations. These results suggest that YOSO not only accelerates the convergence process but also ensures a more stable training path compared to existing sampling methods, highlighting its effectiveness in GNN training

\subsection{Ablation Study}\label{sec: ablation}
In this subsection, we explore how YOSO's total training time and model accuracy vary with different sampling sizes \( M \) and evaluate reconstruction effectiveness by comparing the \( \mathbf{H}^{(L)} \) matrix generated without sampling to the \( \tilde{\mathbf{H}}^{(L)} \) matrix produced by YOSO's sampling-reconstruction process, with the differences visualized with heatmaps.

\begin{figure}[!t]
\begin{minipage}{0.5\textwidth}
    \centerline{\includegraphics[width=\linewidth]{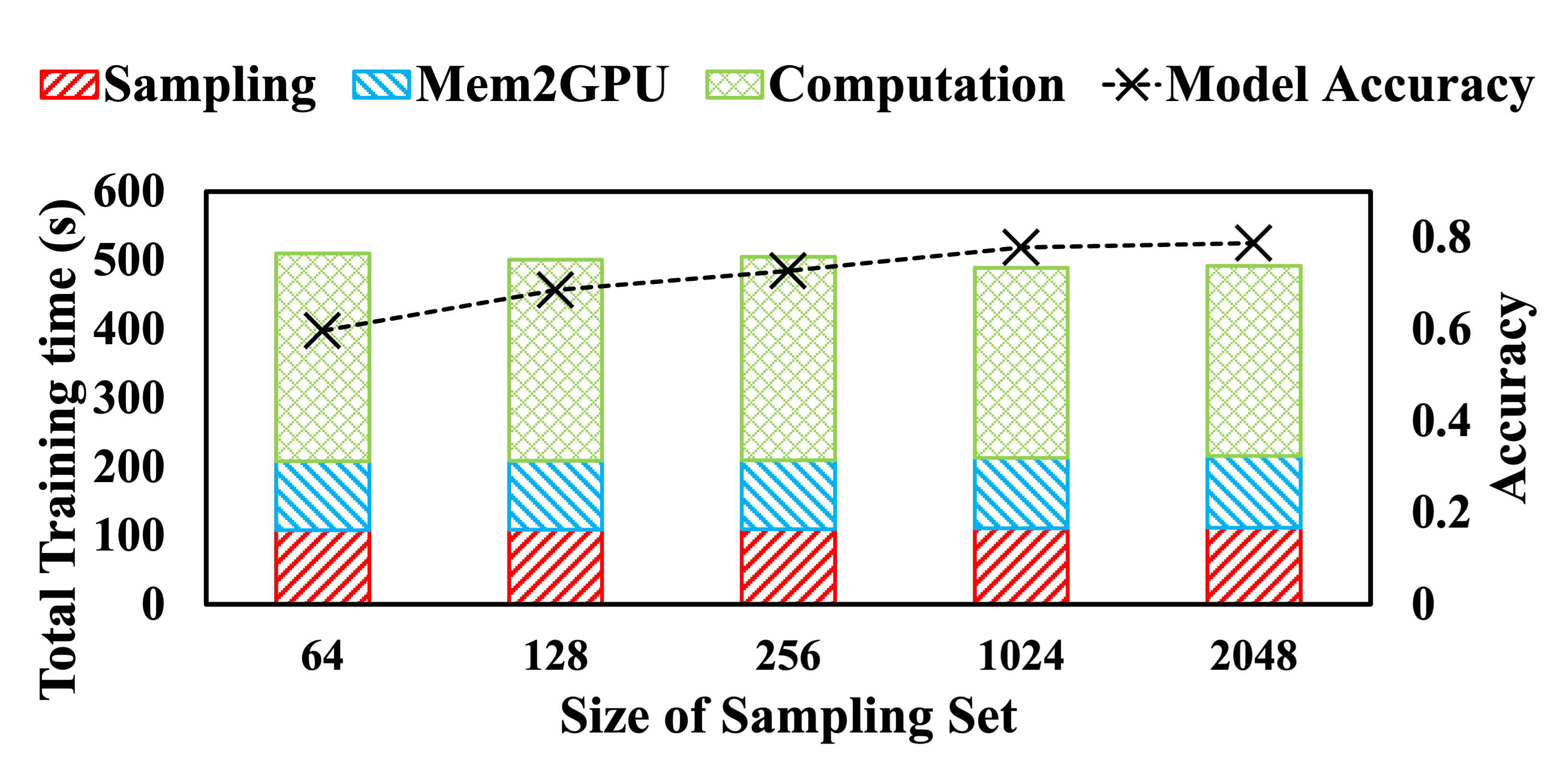}}
    \centerline{(a) ogbn-products}
\end{minipage}
\begin{minipage}{0.5\textwidth}
    \centerline{\includegraphics[width=\linewidth]{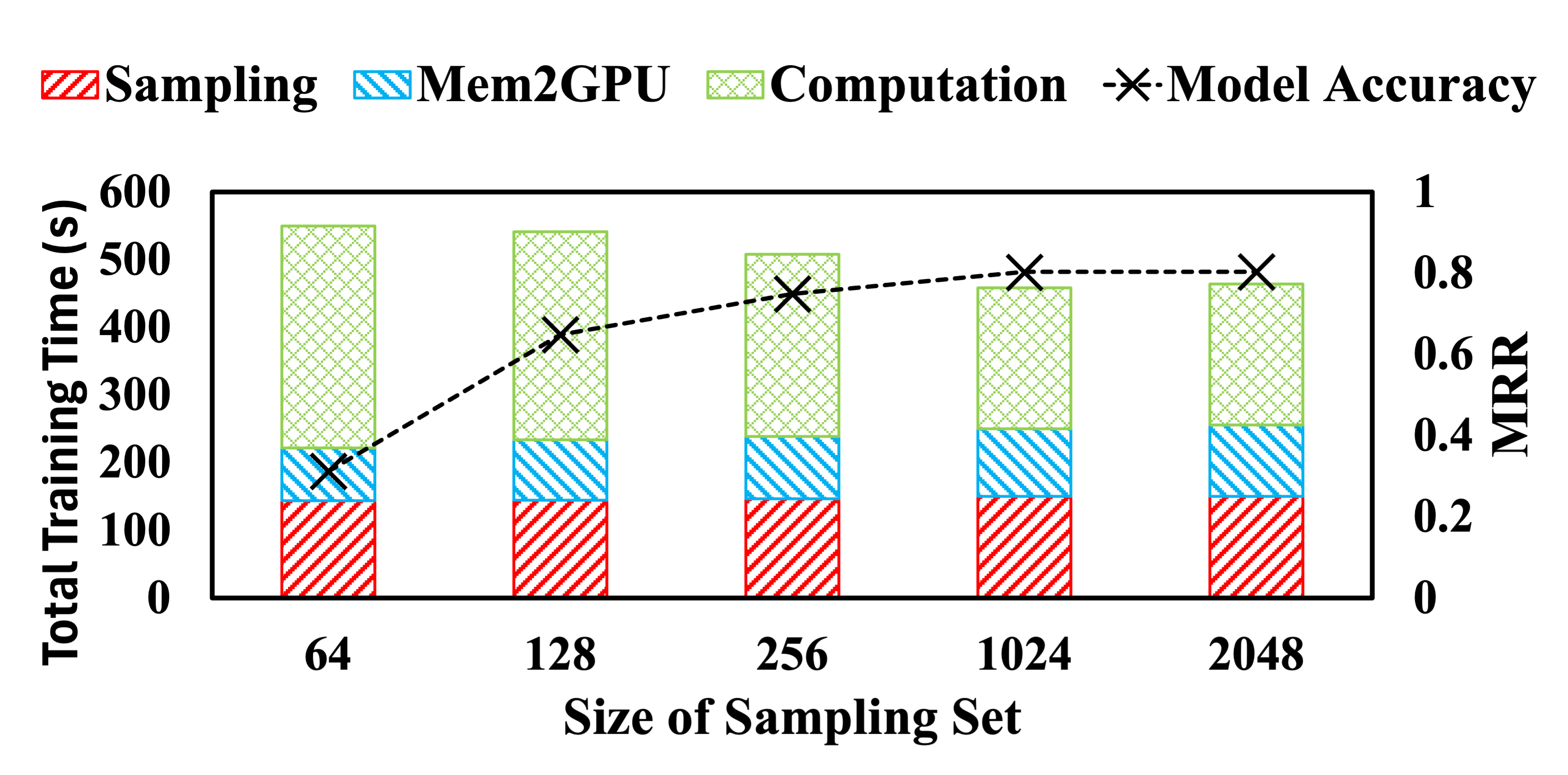}}
    \centerline{(b) ogbl-citation2}
\end{minipage}
\caption{Total training time (including its breakdown) and model accuracy for YOSO with different sampling sizes: (a) for the node classification learning task on the ogbn-products dataset, and (b) for the link prediction learning task on the ogbl-citation2 dataset.}.
\label{fig:varies_M}
\end{figure}

\textbf{Varying sampling size $M$:}
We examine how total training time (including breakdown) and model accuracy vary with \( M \) values, specifically \( M = \{64, 128, 256, 1024, 2048\} \), as shown in Figure~\ref{fig:varies_M}. The results indicate that YOSO's sampling time remains stable across different \( M \), ranging from 107.94 to 111.53 seconds on ogbn-products and 143.56 to 149.65 seconds on ogbl-citation2, showing minimal impact from \( M \). In contrast, as \( M \) decreases, computation time increases, reflecting more iterations needed for convergence (e.g., rising from 275.98s at \( M = 2048 \) to 301.94s at \( M = 64 \) on ogbn-products, with a similar trend on ogbl-citation2). Model accuracy improves with larger \( M \), eventually stabilizing; it rises from 0.597 to 0.7873 on ogbn-products and from 0.312 to 0.8025 on ogbl-citation2. These findings highlight YOSO's efficient sampling and improved accuracy and convergence with larger \( M \).

\begin{figure}[!t]
	\centering
\includegraphics[width=\textwidth]{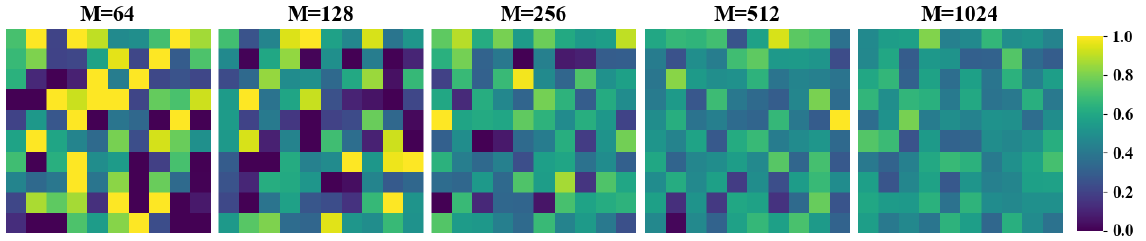}
	\caption{Reconstruction effectiveness visualized via heatmap. Using the ogbn-products dataset, 10 nodes are randomly selected from the training set, and for each node, 10 embedding dimensions are randomly picked. The heatmap shows the absolute differences between original and reconstructed embeddings for these elements. $M$ is the size of the sampling set.}
	\label{fig:heatmap}
\end{figure}
\textbf{Reconstruction effectiveness:} The heatmap in Figure~\ref{fig:heatmap} shows the reconstruction effectiveness for different sampling sizes \( M \). Each $10\times 10$ block represents the absolute difference between reconstructed embeddings from our two-layer GNN sampling and those computed with all neighbors (without sampling). As \( M \) increases, reconstruction accuracy improves, enhancing overall model accuracy. However, beyond a certain point, such as \( M = 512 \) in Figure~\ref{fig:heatmap}, further increases in \( M \) offer diminishing returns in both reconstruction quality and model accuracy. This suggests there is an optimal \( M \) that balances reconstruction quality and computational efficiency.

\section{conclusion}
In this paper, we introduce YOSO (You Only Sample Once), a novel algorithm aimed at significantly enhancing the efficiency of GNN training without sacrificing prediction accuracy. By leveraging a compressed sensing-based sampling and reconstruction framework, YOSO performs node sampling only once at the input layer, followed by a lossless reconstruction at the output layer during each training epoch. Our experimental results demonstrate that YOSO can achieve up to 75\% reduction of existing state-of-the-art methods while achieving accuracy comparable to top-performing baselines.

\newpage
\textbf{Ethics Statement:} In this paper, we present a technique grounded in compressed sensing that addresses the growing computational demands of GNN sampling schemes. Our approach significantly reduces sampling time and overall GNN training duration without compromising model accuracy, thereby enhancing the efficiency of graph neural network training. This improvement holds potential for a wide range of applications, such as recommendation systems and social network analysis, and bioinformatics. We believe that our method contributes positively to the advancement of machine learning research by promoting computational efficiency. Although we do not anticipate any immediate negative ethical implications or societal concerns from our approach, it's important to acknowledge that machine learning technologies, including graph-based methods, have broader impacts. Therefore, responsible implementation is crucial to ensure that such technologies are applied in a manner that promotes fairness and beneficial societal outcomes.

\bibliography{iclr2025_conference}

\begin{thebibliography}{33}
\providecommand{\natexlab}[1]{#1}
\providecommand{\url}[1]{\texttt{#1}}
\expandafter\ifx\csname urlstyle\endcsname\relax
  \providecommand{\doi}[1]{doi: #1}\else
  \providecommand{\doi}{doi: \begingroup \urlstyle{rm}\Url}\fi

\bibitem[Baraniuk et~al.(2008)Baraniuk, Davenport, DeVore, and Wakin]{baraniuk2008simple}
Richard Baraniuk, Mark Davenport, Ronald DeVore, and Michael Wakin.
\newblock A simple proof of the restricted isometry property for random matrices.
\newblock \emph{Constructive approximation}, 28:\penalty0 253--263, 2008.

\bibitem[Bo et~al.(2023)Bo, Wang, Liu, Fang, Li, and Shi]{bo2023survey}
Deyu Bo, Xiao Wang, Yang Liu, Yuan Fang, Yawen Li, and Chuan Shi.
\newblock A survey on spectral graph neural networks.
\newblock \emph{arXiv preprint arXiv:2302.05631}, 2023.

\bibitem[Candes \& Tao(2005)Candes and Tao]{candes2005decoding}
Emmanuel~J Candes and Terence Tao.
\newblock Decoding by linear programming.
\newblock \emph{IEEE transactions on information theory}, 51\penalty0 (12):\penalty0 4203--4215, 2005.

\bibitem[Candes \& Tao(2006)Candes and Tao]{candes2006near}
Emmanuel~J Candes and Terence Tao.
\newblock Near-optimal signal recovery from random projections: Universal encoding strategies?
\newblock \emph{IEEE transactions on information theory}, 52\penalty0 (12):\penalty0 5406--5425, 2006.

\bibitem[Chen et~al.(2017)Chen, Zhu, and Song]{chen2017stochastic}
Jianfei Chen, Jun Zhu, and Le~Song.
\newblock Stochastic training of graph convolutional networks with variance reduction.
\newblock \emph{arXiv preprint arXiv:1710.10568}, 2017.

\bibitem[Chen et~al.(2018)Chen, Ma, and Xiao]{chen2018fastgcn}
Jie Chen, Tengfei Ma, and Cao Xiao.
\newblock Fastgcn: fast learning with graph convolutional networks via importance sampling.
\newblock \emph{arXiv preprint arXiv:1801.10247}, 2018.

\bibitem[Chen et~al.(2015)Chen, Varma, Sandryhaila, and Kova{\v{c}}evi{\'c}]{chen2015discrete}
Siheng Chen, Rohan Varma, Aliaksei Sandryhaila, and Jelena Kova{\v{c}}evi{\'c}.
\newblock Discrete signal processing on graphs: Sampling theory<? pub \_newline=""?
\newblock \emph{IEEE transactions on signal processing}, 63\penalty0 (24):\penalty0 6510--6523, 2015.

\bibitem[Chiang et~al.(2019)Chiang, Liu, Si, Li, Bengio, and Hsieh]{chiang2019cluster}
Wei-Lin Chiang, Xuanqing Liu, Si~Si, Yang Li, Samy Bengio, and Cho-Jui Hsieh.
\newblock Cluster-gcn: An efficient algorithm for training deep and large graph convolutional networks.
\newblock In \emph{Proceedings of the 25th ACM SIGKDD international conference on knowledge discovery \& data mining}, pp.\  257--266, 2019.

\bibitem[Dai et~al.(2018)Dai, Kozareva, Dai, Smola, and Song]{dai2018learning}
Hanjun Dai, Zornitsa Kozareva, Bo~Dai, Alex Smola, and Le~Song.
\newblock Learning steady-states of iterative algorithms over graphs.
\newblock In \emph{International conference on machine learning}, pp.\  1106--1114. PMLR, 2018.

\bibitem[Dunteman(1989)]{dunteman1989principal}
George~H Dunteman.
\newblock \emph{Principal components analysis}, volume~69.
\newblock Sage, 1989.

\bibitem[Guo \& Wang(2020)Guo and Wang]{guo2020deep}
Zhiwei Guo and Heng Wang.
\newblock A deep graph neural network-based mechanism for social recommendations.
\newblock \emph{IEEE Transactions on Industrial Informatics}, 17\penalty0 (4):\penalty0 2776--2783, 2020.

\bibitem[Hamilton et~al.(2017)Hamilton, Ying, and Leskovec]{hamilton2017inductive}
Will Hamilton, Zhitao Ying, and Jure Leskovec.
\newblock Inductive representation learning on large graphs.
\newblock \emph{Advances in neural information processing systems}, 30, 2017.

\bibitem[Hu et~al.(2020)Hu, Fey, Zitnik, Dong, Ren, Liu, Catasta, and Leskovec]{hu2020open}
Weihua Hu, Matthias Fey, Marinka Zitnik, Yuxiao Dong, Hongyu Ren, Bowen Liu, Michele Catasta, and Jure Leskovec.
\newblock Open graph benchmark: Datasets for machine learning on graphs.
\newblock \emph{Advances in neural information processing systems}, 33:\penalty0 22118--22133, 2020.

\bibitem[Huang et~al.(2018)Huang, Zhang, Rong, and Huang]{huang2018adaptive}
Wenbing Huang, Tong Zhang, Yu~Rong, and Junzhou Huang.
\newblock Adaptive sampling towards fast graph representation learning.
\newblock \emph{Advances in neural information processing systems}, 31, 2018.

\bibitem[Isufi et~al.(2024)Isufi, Gama, Shuman, and Segarra]{isufi2024graph}
Elvin Isufi, Fernando Gama, David~I Shuman, and Santiago Segarra.
\newblock Graph filters for signal processing and machine learning on graphs.
\newblock \emph{IEEE Transactions on Signal Processing}, 2024.

\bibitem[Jin et~al.(2020)Jin, Ma, Liu, Tang, Wang, and Tang]{jin2020graph}
Wei Jin, Yao Ma, Xiaorui Liu, Xianfeng Tang, Suhang Wang, and Jiliang Tang.
\newblock Graph structure learning for robust graph neural networks.
\newblock In \emph{Proceedings of the 26th ACM SIGKDD international conference on knowledge discovery \& data mining}, pp.\  66--74, 2020.

\bibitem[Khatua et~al.(2023)Khatua, Mailthody, Taleka, Ma, Song, and Hwu]{khatua2023igb}
Arpandeep Khatua, Vikram~Sharma Mailthody, Bhagyashree Taleka, Tengfei Ma, Xiang Song, and Wen-mei Hwu.
\newblock Igb: Addressing the gaps in labeling, features, heterogeneity, and size of public graph datasets for deep learning research.
\newblock In \emph{Proceedings of the 29th ACM SIGKDD Conference on Knowledge Discovery and Data Mining}, pp.\  4284--4295, 2023.

\bibitem[Kipf \& Welling(2016)Kipf and Welling]{kipf2016semi}
Thomas~N Kipf and Max Welling.
\newblock Semi-supervised classification with graph convolutional networks.
\newblock \emph{arXiv preprint arXiv:1609.02907}, 2016.

\bibitem[Koochakzadeh et~al.(2016)Koochakzadeh, Miran, Samangouei, and Rotkowitz]{koochakzadeh2016nonnegative}
Ali Koochakzadeh, Sina Miran, Pouya Samangouei, and Michael~C Rotkowitz.
\newblock Nonnegative matrix factorization by optimization on the stiefel manifold with svd initialization.
\newblock In \emph{2016 54th Annual Allerton Conference on Communication, Control, and Computing (Allerton)}, pp.\  1068--1073. IEEE, 2016.

\bibitem[Liu et~al.(2021{\natexlab{a}})Liu, Wu, Zhuang, Lu, Dou, and Xiong]{liu2021community}
Hao Liu, Qiyu Wu, Fuzhen Zhuang, Xinjiang Lu, Dejing Dou, and Hui Xiong.
\newblock Community-aware multi-task transportation demand prediction.
\newblock In \emph{Proceedings of the AAAI Conference on Artificial Intelligence}, volume~35, pp.\  320--327, 2021{\natexlab{a}}.

\bibitem[Liu et~al.(2021{\natexlab{b}})Liu, Yan, Deng, Li, Ye, and Fan]{liu2021sampling}
Xin Liu, Mingyu Yan, Lei Deng, Guoqi Li, Xiaochun Ye, and Dongrui Fan.
\newblock Sampling methods for efficient training of graph convolutional networks: A survey.
\newblock \emph{IEEE/CAA Journal of Automatica Sinica}, 9\penalty0 (2):\penalty0 205--234, 2021{\natexlab{b}}.

\bibitem[Liu et~al.(2018)Liu, Gao, Gao, and Shao]{liu2018l_}
Yang Liu, Quanxue Gao, Xinbo Gao, and Ling Shao.
\newblock $l_{2,1}$-norm discriminant manifold learning.
\newblock \emph{IEEE Access}, 6:\penalty0 40723--40734, 2018.

\bibitem[Maleki(2010)]{maleki2010approximate}
Arian Maleki.
\newblock \emph{Approximate message passing algorithms for compressed sensing}.
\newblock PhD thesis, Stanford University, 2010.

\bibitem[Mikolov et~al.(2013)Mikolov, Sutskever, Chen, Corrado, and Dean]{mikolov2013distributed}
Tomas Mikolov, Ilya Sutskever, Kai Chen, Greg~S Corrado, and Jeff Dean.
\newblock Distributed representations of words and phrases and their compositionality.
\newblock \emph{Advances in neural information processing systems}, 26, 2013.

\bibitem[Pennington et~al.(2014)Pennington, Socher, and Manning]{pennington2014glove}
Jeffrey Pennington, Richard Socher, and Christopher~D Manning.
\newblock Glove: Global vectors for word representation.
\newblock In \emph{Proceedings of the 2014 conference on empirical methods in natural language processing (EMNLP)}, pp.\  1532--1543, 2014.

\bibitem[Puy et~al.(2018)Puy, Tremblay, Gribonval, and Vandergheynst]{puy2018random}
Gilles Puy, Nicolas Tremblay, R{\'e}mi Gribonval, and Pierre Vandergheynst.
\newblock Random sampling of bandlimited signals on graphs.
\newblock \emph{Applied and Computational Harmonic Analysis}, 44\penalty0 (2):\penalty0 446--475, 2018.

\bibitem[R{\'e}au et~al.(2023)R{\'e}au, Renaud, Xue, and Bonvin]{reau2023deeprank}
Manon R{\'e}au, Nicolas Renaud, Li~C Xue, and Alexandre~MJJ Bonvin.
\newblock Deeprank-gnn: a graph neural network framework to learn patterns in protein--protein interfaces.
\newblock \emph{Bioinformatics}, 39\penalty0 (1):\penalty0 btac759, 2023.

\bibitem[Tsitsvero et~al.(2016)Tsitsvero, Barbarossa, and Di~Lorenzo]{tsitsvero2016signals}
Mikhail Tsitsvero, Sergio Barbarossa, and Paolo Di~Lorenzo.
\newblock Signals on graphs: Uncertainty principle and sampling.
\newblock \emph{IEEE Transactions on Signal Processing}, 64\penalty0 (18):\penalty0 4845--4860, 2016.

\bibitem[Veli{\v{c}}kovi{\'c} et~al.(2017)Veli{\v{c}}kovi{\'c}, Cucurull, Casanova, Romero, Lio, and Bengio]{velivckovic2017graph}
Petar Veli{\v{c}}kovi{\'c}, Guillem Cucurull, Arantxa Casanova, Adriana Romero, Pietro Lio, and Yoshua Bengio.
\newblock Graph attention networks.
\newblock \emph{arXiv preprint arXiv:1710.10903}, 2017.

\bibitem[Wang et~al.(2020)Wang, Shen, Huang, Wu, Dong, and Kanakia]{wang2020microsoft}
Kuansan Wang, Zhihong Shen, Chiyuan Huang, Chieh-Han Wu, Yuxiao Dong, and Anshul Kanakia.
\newblock Microsoft academic graph: When experts are not enough.
\newblock \emph{Quantitative Science Studies}, 1\penalty0 (1):\penalty0 396--413, 2020.

\bibitem[Ying et~al.(2018)Ying, He, Chen, Eksombatchai, Hamilton, and Leskovec]{ying2018graph}
Rex Ying, Ruining He, Kaifeng Chen, Pong Eksombatchai, William~L Hamilton, and Jure Leskovec.
\newblock Graph convolutional neural networks for web-scale recommender systems.
\newblock In \emph{Proceedings of the 24th ACM SIGKDD international conference on knowledge discovery \& data mining}, pp.\  974--983, 2018.

\bibitem[Zeng et~al.(2019)Zeng, Zhou, Srivastava, Kannan, and Prasanna]{zeng2019graphsaint}
Hanqing Zeng, Hongkuan Zhou, Ajitesh Srivastava, Rajgopal Kannan, and Viktor Prasanna.
\newblock Graphsaint: Graph sampling based inductive learning method.
\newblock \emph{arXiv preprint arXiv:1907.04931}, 2019.

\bibitem[Zou et~al.(2019)Zou, Hu, Wang, Jiang, Sun, and Gu]{zou2019layer}
Difan Zou, Ziniu Hu, Yewen Wang, Song Jiang, Yizhou Sun, and Quanquan Gu.
\newblock Layer-dependent importance sampling for training deep and large graph convolutional networks.
\newblock \emph{Advances in neural information processing systems}, 32, 2019.

\end{thebibliography}
\bibliographystyle{iclr2025_conference}

\newpage
\appendix
\section{Details about experiments}
\subsection{Hardware and Software Configuration}\label{app: hard/software}
We evaluate all baselines and our design on a Linux Desktop running Ubuntu 18.04.6 LTS, equipped with an NVIDIA GTX 1060Ti (6GB memory) using CUDA version 11.8 and PyTorch version 2.0.0. The system features a AMD Ryzen 5 5500 CPU with 64 GB DDR4 RAM, and the Python version used is 3.9.0.

\subsection{Datasets}\label{app: datasets}
\textbf{Data splitting}: We adopt strategies consistent with previous works~\citep{hamilton2017inductive, hu2020open}. Specifically, for the Reddit dataset, we follow the data splitting used in GraphSage~\citep{hamilton2017inductive}, and for the OGB series (ogbn and ogbl), we maintain the splitting described in~\citep{hu2020open}. 

The basic summary information of the datasets we use is provided in Table~\ref{tab:dataset}, and detailed descriptions are as follows:\\
\textbf{ogbn-arxiv}: This dataset is a directed citation network of Computer Science (CS) arXiv papers from the Microsoft Academic Graph (MAG)~\citep{wang2020microsoft}. Each node represents a paper, with directed edges indicating citations. The task is to classify unlabeled papers into primary categories using labeled papers and node features, which are derived by averaging word2vec embeddings~\citep{mikolov2013distributed} of paper titles and abstracts.\\
\textbf{Reddit}: Originally from GraphSage~\citep{hamilton2017inductive}, this Reddit dataset is a post-to-post graph where each node represents a post, and edges indicate shared user comments. The task is to classify posts into communities using GloVe word vectors~\citep{pennington2014glove} from post titles and comments, along with features such as post scores and comment counts.\\
\textbf{ogbn-products}: This undirected, unweighted graph represents an Amazon product co-purchasing network, where nodes are products and edges indicate frequent co-purchases. Node features are derived from bag-of-words features of product descriptions, reduced to 100 dimensions via Principal Component Analysis~\citep{dunteman1989principal}.\\
\textbf{ogbl-ppa}: This undirected, unweighted graph has nodes representing proteins from 58 species, with edges indicating biologically meaningful associations. Each node features a 58-dimensional one-hot vector for the protein's species. The task is to predict new association edges, evaluated by ranking positive test edges over negative ones.\\
\textbf{ogbl-citation2}: This dataset is a directed graph representing a citation network among a subset of papers from Microsoft Academic Graph (MAG), similar to ogbn-arxiv. For each source paper, two references are randomly removed, and the task is to rank these missing references above 1,000 randomly selected negative references, which are sampled from all papers not cited by the source paper.

\begin{table}[]
\centering
\caption{Statistics and metrics of the dataset}
\label{tab:dataset}
\begin{tabular}{|cc|c|c|c|c|}
\hline
\multicolumn{2}{|c|}{\textbf{Dataset}} &
  \textbf{\#Node} &
  \textbf{\#Edge} &
  \textbf{\#Dim.} &
  \textbf{Metric} \\ \hline
\multicolumn{1}{|c|}{\multirow{3}{*}{\textbf{\begin{tabular}[c]{@{}c@{}}Node Property \\ Prediction\end{tabular}}}} &
  ogbn-arxiv &
  169,343 &
  1,166,243 &
  128 &
  Accuracy \\ \cline{2-6} 
\multicolumn{1}{|c|}{} &
  Reddit &
  232,965 &
  11,606,919 &
  602 &
  Mirco-F1 \\ \cline{2-6} 
\multicolumn{1}{|c|}{} &
  ogbn-products &
  2,449,029 &
  61,859,140 &
  100 &
  Accuracy \\ \hline
\multicolumn{1}{|c|}{\multirow{2}{*}{\textbf{\begin{tabular}[c]{@{}c@{}}Link Property \\ Prediction\end{tabular}}}} &
  ogbl-ppa &
  576,289 &
  30,326,273 &
  128 &
  Hits@100 \\ \cline{2-6} 
\multicolumn{1}{|c|}{} &
  ogbl-citation2 &
  2,927,963 &
  30,561,187 &
  128 &
  MRR \\ \hline
\end{tabular}
\end{table}

\subsection{Baselines and Implementation}\label{app: baselines}
\begin{table}[]
\centering
\caption{Baselines and their public available source code link}
\label{tab:basesline_link}
\begin{tabular}{|c|c|}
\hline
\textbf{Method} & \textbf{Available Link}                                   \\ \hline
GraphSage       & https://github.com/williamleif/graphsage-simple           \\ \hline
VR-GCN          & https://github.com/THUDM/cogdl/tree/master/examples/VRGCN \\ \hline
FastGCN         & https://github.com/gmancino/fastgcn-pytorch               \\ \hline
AS-GCN          & https://github.com/Gkunnan97/FastGCN\_pytorch             \\ \hline
LADIES          & https://github.com/acbull/LADIES                         \\ \hline
Cluster-GCN     & https://github.com/benedekrozemberczki/ClusterGCN         \\ \hline
GraphSAINT      & https://github.com/GraphSAINT/GraphSAINT                  \\ \hline
\end{tabular}
\end{table}
Table~\ref{tab:basesline_link} presents the baselines used in this paper along with their publicly available source code links. Since some baselines were not originally implemented in PyTorch, we standardized the framework for fair comparison. If a PyTorch version involved the original authors, we selected that source code (e.g., FastGCN~\citep{chen2018fastgcn}). Otherwise, we chose the most popular implementation based on the number of stars. Notably, the repository linked for AS-GCN~\citep{huang2018adaptive} in the table includes implementations of both FastGCN and AS-GCN, but we only used the AS-GCN version, while the FastGCN implementation was taken from the source listed in the table.\newline
\textbf{YOSO's Implementation}: The base code of YOSO\footnote{https://anonymous.4open.science/r/YOSO-B49B} is built on GCN~\citep{kipf2016semi}, with the link available at https://github.com/tkipf/pygcn. The sampling stage in YOSO occurs on the CPU and main memory since it involves calculations related to the entire feature matrix and the regularized Laplacian matrix. After sampling, the relevant data is migrated to GPU memory for computation. Throughout the training process, multiple data exchanges occur between main memory and GPU memory, such as in link prediction tasks where node embeddings need to be updated.\newline
\textbf{Modification:} All baselines support updating node embeddings and performing node classification tasks. For node classification, if a baseline did not originally use the cross-entropy loss function, we adjusted it to adopt this loss function. For the link prediction task, the following loss function is applied:
\[
    \mathcal{L} = \frac{1}{N^{+}}\sum_{(i,j)\in E^{+}}\left(1 - \frac{\mathbf{h}_{i}^{(L)} \cdot \mathbf{h}_{j}^{(L)}}{\Vert \mathbf{h}_{i}^{(L)} \Vert \Vert \mathbf{h}_{j}^{(L)} \Vert}\right) + \frac{1}{N^{-}}\sum_{(i,j)\in E^{-}}\text{max}\left(0, \gamma - \left(1 - \frac{\mathbf{h}_{i}^{(L)} \cdot \mathbf{h}_{j}^{(L)}}{\Vert \mathbf{h}_{i}^{(L)} \Vert \Vert \mathbf{h}_{j}^{(L)} \Vert}\right)\right)
\]
where \( N^{+} \) and \( N^{-} \) represent the number of positive and negative samples, respectively, and \( E^{+} \) and \( E^{-} \) denote the sets of positive and negative edges. The parameter \( \gamma \) is a hyperparameter, set to 0.5 in this study. As the ogbl-ppa and ogbl-citation2 datasets provide corresponding negative edges by default, we used these pre-defined negative edges for our calculations.

\subsection{Hyper-parameter Setting}\label{app: hyperparameter}
The hyperparameter settings for both YOSO and the baselines are provided in Table~\ref{tab:hyper_node_setting} and Table~\ref{tab:hyper_link_setting} for node classification and link prediction datasets, respectively. All experiments were conducted using a two-layer GCN with official configurations. When certain parameters were not clearly specified in some papers, we fine-tuned them for optimal accuracy. The recorded hyperparameters include the sampling size (per node/layer/subgraph), the optimizer, and the learning rate. For YOSO, the sampling size is denoted as \( M \); for example, on the ogbl-ppa dataset (Table~\ref{tab:hyper_link_setting}), \( M = 128 \).
\begin{table}[]
\centering
\caption{Node classification hyperparamter setting for baselines and YOSO on different datasets.}
\label{tab:hyper_node_setting}
\begin{tabular}{|c|c|c|c|}
\hline
                       & ogbn-arxiv          & Reddit               & ogbn-products        \\ \hline
GraphSage              & 25\&10 / Adam / 0.7 & 25\&10 / Adam / 0.01 & 50\&20 / Adam / 0.01 \\ \hline
VR-GCN                 & 8 / Adam / 0.01     & 16 / Adam / 0.01     & 32 / Adam / 0.01     \\ \hline
FastGCN                & 64 / Adam / 0.01    & 128 / Adam / 0.001   & 256 / Adam / 0.001   \\ \hline
AS-GCN                 & 128 / Adam / 0.001  & 512 / Adam / 0.01    & 1000 / Adam / 0.01   \\ \hline
LADIES                 & 64 / Adam / 0.001   & 128 / Adam / 0.001   & 256 / Adam / 0.001   \\ \hline
Cluster-GCN            & - / Adam / 0.01     & - / Adam / 0.005     & - / Adam / 0.005     \\ \hline
GraphSAINT-EG        & 300 / Adam / 0.01   & 600 / Adam / 0.01    & 4000 / Adam / 0.01   \\ \hline
GraphSAINT-RW & 4000 / Adam / 0.01  & 8000 / Adam / 0.01   & 10000 / Adam / 0.01  \\ \hline
YOSO                   & 128 / Adam / 0.01   & 256 / Adam / 0.01    & 512 / Adam / 0.01    \\ \hline
\end{tabular}
\end{table}
\begin{table}[]
\centering
\caption{Link prediction hyperparamter setting for baselines and YOSO on different datasets.}
\label{tab:hyper_link_setting}
\begin{tabular}{|c|c|c|}
\hline
              & ogbl-ppa            & ogbl-citation2       \\ \hline
GraphSage     & 25\&10 / Adam / 0.7 & 50\&20 / Adam / 0.01 \\ \hline
VR-GCN        & 8 / Adam / 0.01     & 32 / Adam / 0.01     \\ \hline
FastGCN       & 64 / Adam / 0.01    & 256 / Adam / 0.001   \\ \hline
AS-GCN        & 128 / Adam / 0.001  & 1000 / Adam / 0.01   \\ \hline
LADIES        & 64 / Adam / 0.001   & 256 / Adam / 0.001   \\ \hline
Cluster-GCN   & - / Adam / 0.01     & - / Adam / 0.005     \\ \hline
GraphSAINT-EG & 300 / Adam / 0.01   & 4000 / Adam / 0.01   \\ \hline
GraphSAINT-RW & 4000 / Adam / 0.01  & 10000 / Adam / 0.01  \\ \hline
YOSO          & 128 / Adam / 0.01   & 512 / Adam / 0.01    \\ \hline
\end{tabular}
\end{table}

\section{Computation and Proof}\label{app:proof}
\subsection{Gradient Computation}\label{app:gradient}
\subsubsection{Computation of $\nabla_{\Theta}\mathcal{L}$:}\label{app:theta}
$\nabla_{\Theta}\mathcal{L} = \alpha\nabla_{\Theta}\mathcal{L}_{recon}+\beta\nabla_{\Theta}\mathcal{L}_{GNN}^{\Theta}(\mathbf{Z}) = \frac{\partial \mathcal{L}_{recon}}{\partial \mathbf{Z}}\cdot \frac{\partial \mathbf{Z}}{\partial \Theta}+\frac{\partial\mathcal{L}_{GNN}^{\Theta}(\mathbf{Z})}{\partial \mathbf{Z}}\cdot \frac{\partial \mathbf{Z}}{\partial \Theta}$
\begin{itemize}[leftmargin=*]
    \item $\frac{\partial \mathcal{L}_{recon}}{\partial \mathbf{Z}}= (\mathbf{Z} - \mathbf{\Phi}\mathbf{U}\hat{\mathbf{H}}^{(L)})$
    \item Consider the $g^{(L)}$ which is the gradient at the output layer, and we have $g^{(L)} = \frac{\partial\mathcal{L}_{recon}}{\partial\mathbf{Z}}\odot \sigma^{'}(\mathbf{S}^{(L)})$ where $\odot$ denotes element-wise multiplication, $\sigma^{'}(\mathbf{S}^{(L)})$ is the derivation of the activation function at layer $L$ and $\mathbf{S}^{(L)}$ is the pre-activation input at layer $L$. Therefore, for $l=L, L-1,..., 1$, we have $g^{(l-1)} = \nabla_{\mathbf{W}^{(L)}}\mathcal{L}_{recon}\odot\sigma^{'}(\mathbf{S}^{(l-1)}) = (\mathbf{\Phi}\hat{\mathbf{A}}\mathbf{W}^{(l)})^Tg^{(l)}\odot\sigma^{'}(\mathbf{S}^{(l-1)})$. By iteratively executing this process, we can obtain $\frac{\partial \mathbf{Z}}{\partial \Theta}$
    \item $\frac{\partial\mathcal{L}_{GNN}^{\Theta}}{\partial\mathbf{Z}}$ depends on the specific loss function used.
\end{itemize}
\subsubsection{Computation of $\nabla_{\mathbf{U}}\mathcal{L}$}
$\nabla_{\mathbf{U}}\mathcal{L} = \alpha\nabla_{\mathbf{U}}\mathcal{L}_{recon}+\beta\nabla_{\mathbf{U}}\mathcal{L}_{GNN}^{\Theta}(\mathbf{Z}) = \alpha\nabla_{\mathbf{U}}\mathcal{L}_{recon} + \beta(\frac{\partial\mathcal{L}_{GNN}^{\Theta}(\mathbf{Z})}{\partial \mathbf{Z}}\cdot \frac{\partial \mathbf{Z}}{\partial \mathbf{U}})$
\begin{itemize}[leftmargin=*]
    \item $\nabla_{\mathbf{U}}\mathcal{L}_{recon} = -\mathbf{\Phi}^{T}(\mathbf{Z} - \mathbf{\Phi}\mathbf{U}\hat{\mathbf{H}}^{(L)})(\hat{\mathbf{H}}^{(L)})^{T}$
    \item As in Section~\ref{app:theta}, $\frac{\partial\mathcal{L}_{GNN}^{\Theta}}{\partial\mathbf{Z}}$ depends on specific loss function and easy to compute.
    \item For $\frac{\partial \mathbf{Z}}{\partial \mathbf{U}}$, it need to be computed recursively. Since $\mathbf{T}^{(0)}=\mathbf{\Phi}\mathbf{U}\hat{\mathbf{X}}$, $\frac{\partial \mathbf{T}^{(0)}}{\partial \mathbf{U}} = \mathbf{\Phi}\hat{\mathbf{X}}$. The gradient propagates from $\mathbf{Z}$ back to $\mathbf{U}$: $\nabla_{\mathbf{U}}\mathcal{L}_{GNN}^{\Theta}(\mathbf{Z}) = (\frac{\partial\mathcal{L}_{GNN}^{\Theta}(\mathbf{Z})}{\partial \mathbf{Z}}\cdot \frac{\partial \mathbf{Z}}{\partial\mathbf{T}^{(L-1)}}\cdots \frac{\partial \mathbf{T}^{(1)}}{\partial\mathbf{T}^{(0)}}\cdot \frac{\partial \mathbf{T}^{(0)}}{\partial\mathbf{U}})$. As we know that $\mathbf{T}^{(l)} = \sigma(\mathbf{S}^{(l)})$ and $\mathbf{S}^{(l)} = \mathbf{\Phi}\hat{\mathbf{A}}\mathbf{W}^{(l)}\mathbf{T}^{(l-1)}$, therefore $\frac{\partial\mathbf{T}^{(l-1)}}{\partial \mathbf{T}^{(l-1)}} = (\mathbf{\Phi}\hat{\mathbf{A}}\mathbf{W}^{(l)})^T\text{diag}(\sigma^{'}(\mathbf{S}^{(l)}))$
\end{itemize}
\subsubsection{Computation of $\nabla_{\hat{\mathbf{H}}^{(L)}}\mathcal{L}$}
$\nabla_{\hat{\mathbf{H}}^{(L)}}\mathcal{L} = \alpha\nabla_{\hat{\mathbf{H}}^{(L)}}\mathcal{L}_{recon} = -\mathbf{U}^{T}\mathbf{\Phi}^{T}(\mathbf{Z} - \mathbf{\Phi}\mathbf{U}\hat{\mathbf{H}}^{(L)})+\lambda\partial\Vert\hat{\mathbf{H}}^{(L)} \Vert_{2,1}$ where $\partial\Vert\hat{\mathbf{H}}^{(L)} \Vert_{2,1}$ is the subgradient of the $l_{2,1}$ norm and computed as $(\partial\Vert\hat{\mathbf{H}}^{(L)} \Vert_{2,1})_i = \frac{\hat{\mathbf{H}}^{(L)}_{i,:}}{\Vert \hat{\mathbf{H}}^{(L)}_{i,:} \Vert_2}$ if and only if $\hat{\mathbf{H}}^{(L)}_{i,:} \neq 0$, otherwise, $(\partial\Vert\hat{\mathbf{H}}^{(L)} \Vert_{2,1})_i=0$

\subsection{Full Rank of $\mathbf{\Phi}$}\label{app:full_rank}
\textbf{Theorem 1:} Let $\hat{\mathbf{S}} \in \mathbb{R}^{M \times N}$ be a binary sampling matrix derived from the graph's structure, where each entry $\hat{\mathbf{S}}_{i,j} \in \{0,1\}$ and each row has at least one non-zero entry. Let $\mathbf{\Sigma} \in \mathbb{R}^{M \times N}$ be a random matrix with entries drawn independently from a continuous probability distribution. Define $\mathbf{\Phi} = \hat{\mathbf{S}} \otimes \mathbf{\Sigma}$, where $\otimes$ denotes element-wise multiplication. Then, with probability $1$, the matrix $\mathbf{\Phi}$ has full row rank $M$.\newline
\textbf{Proof:} First, we know that the structure of $\mathbf{\Phi}$ satisfies the following conditions: 
\begin{itemize}[leftmargin=*]
    \item Each entry of $\mathbf{\Phi}$ is given by $\mathbf{\Phi}_{i,j} = \hat{\mathbf{S}}_{i,j} \cdot \mathbf{\Sigma}_{i,j}$.
    \item The $i$-th row of $\mathbf{\Phi}$ is $\mathbf{\Phi}_{i,:} = \hat{\mathbf{S}}_{i,:} \otimes \mathbf{\Sigma}_{i,:}$.
    \item Non-zero entries in $\mathbf{\Phi}_{i,:}$ correspond to positions where $\hat{\mathbf{S}}_{i,j} = 1$.
\end{itemize}
Assume there exist scalars \( c_1, c_2, \dots, c_M \), not all zero, such that \( \sum_{i=1}^{M} c_i \mathbf{\Phi}_{i,:} = \mathbf{0} \). This implies that for each \( j = 1, \dots, N \), we have \( \sum_{i=1}^{M} c_i \hat{\mathbf{S}}_{i,j} \mathbf{\Sigma}_{i,j} = 0 \). Let \( I_j = \{ i \mid \hat{\mathbf{S}}_{i,j} = 1 \} \); then \( \sum_{i \in I_j} c_i \mathbf{\Sigma}_{i,j} = 0 \). 

Since the \( \mathbf{\Sigma}_{i,j} \) values are independently drawn from continuous distributions, the probability that this equation holds for any non-zero set of \( \{ c_i \} \) is zero unless all \( c_i \) in \( I_j \) are zero. Therefore, for the equation to be valid, \( c_i = 0 \) for all \( i \) where \( \hat{\mathbf{S}}_{i,j} = 1 \). 

As each row \( i \) contains at least one entry with \( \hat{\mathbf{S}}_{i,j} = 1 \), it follows that \( c_i = 0 \) for all \( i \). This contradicts the assumption that not all \( c_i \) are zero. Hence, the only solution is \( c_i = 0 \) for all \( i \), indicating that the rows of \( \mathbf{\Phi} \) are linearly independent with probability 1. Thus, \( \text{rank}(\mathbf{\Phi}) = M \) with probability 1.

\subsection{Sampling Matrix $\mathbf{\Phi}$ and RIP}\label{app:Phi_RIP}
\textbf{Theorem 2:} Let $\hat{\mathbf{S}} \in \mathbb{R}^{M \times N}$ be a selection matrix derived from the graph's structure, where each entry $\hat{\mathbf{S}}_{i,j} \in \{0,1\}$ indicates whether node $j$ is included in the $i$-th measurement. Let $\mathbf{\Sigma} \in \mathbb{R}^{M \times N}$ be a matrix whose entries $\mathbf{\Sigma}_{i,j}$ are independent sub-Gaussian random variables with mean zero and variance $\frac{1}{g(j)}$, where $g(j) > 0$. Define the sampling matrix $\mathbf{\Phi} = \hat{\mathbf{S}} \otimes \mathbf{\Sigma}$, where $\otimes$ denotes element-wise multiplication. Then, for any $0 < \delta_k < 1$, there exists a constant $c > 0$ such that if $M \geq c \cdot k \log\left( \frac{N}{k} \right)$, then with probability at least $1 - e^{-cM}$, the matrix $\mathbf{\Phi} \mathbf{U}$ satisfies the Restricted Isometry Property (RIP) of order $k$ with constant $\delta_k$; that is, for all $\hat{\mathbf{H}} \in \mathbb{R}^{N \times d}$ with $\| \hat{\mathbf{H}} \|_{0, \text{row}} \leq k$,
\[
(1 - \delta_k) \| \hat{\mathbf{H}} \|_F^2 \leq \| \mathbf{\Phi} \mathbf{U} \hat{\mathbf{H}} \|_F^2 \leq (1 + \delta_k) \| \hat{\mathbf{H}} \|_F^2.
\]\newline
\textbf{Proof:} To demonstrate that \( \mathbf{\Phi} \mathbf{U} \) satisfies the Restricted Isometry Property (RIP) of order \( k \) with high probability, we consider \( \mathbf{\Phi} \mathbf{U} \hat{\mathbf{H}} = (\hat{\mathbf{S}} \otimes \mathbf{\Sigma}) \mathbf{U} \hat{\mathbf{H}} \). For each row \( i \) and column \( r \), the entry \( (\mathbf{\Phi} \mathbf{U} \hat{\mathbf{H}})_{i,r} \) can be expressed as \( \sum_{j=1}^N \hat{\mathbf{S}}_{i,j} \mathbf{\Sigma}_{i,j} (\mathbf{U} \hat{\mathbf{H}})_{j,r} \). This sum only involves terms where \( \hat{\mathbf{S}}_{i,j} = 1 \). Therefore, \( (\mathbf{\Phi} \mathbf{U} \hat{\mathbf{H}})_{i,r} = \sum_{j \in \mathcal{S}_i} \mathbf{\Sigma}_{i,j} (\mathbf{U} \hat{\mathbf{H}})_{j,r} \), where \( \mathcal{S}_i = \{ j \mid \hat{\mathbf{S}}_{i,j} = 1 \} \).

The variables \( \mathbf{\Sigma}_{i,j} \) are independent sub-Gaussian random variables with mean zero and variance \( \frac{1}{g(j)} \). Therefore, the expectation of \( \| \mathbf{\Phi} \mathbf{U} \hat{\mathbf{H}} \|_F^2 \) can be computed as follows:
\[
\mathbb{E} \left[ \| \mathbf{\Phi} \mathbf{U} \hat{\mathbf{H}} \|_F^2 \right] = \sum_{i=1}^M \sum_{r=1}^d \mathbb{E} \left[ \left( \sum_{j \in \mathcal{S}_i} \mathbf{\Sigma}_{i,j} (\mathbf{U} \hat{\mathbf{H}})_{j,r} \right)^2 \right]
\]
Expanding this and leveraging the independence of \( \mathbf{\Sigma}_{i,j} \), we have:
\[
\mathbb{E} \left[ \left( \sum_{j \in \mathcal{S}_i} \mathbf{\Sigma}_{i,j} (\mathbf{U} \hat{\mathbf{H}})_{j,r} \right)^2 \right] = \sum_{j \in \mathcal{S}_i} \mathbb{E} \left[ \mathbf{\Sigma}_{i,j}^2 \right] \left( (\mathbf{U} \hat{\mathbf{H}})_{j,r} \right)^2
\]
Since \( \mathbb{E}[\mathbf{\Sigma}_{i,j}^2] = \frac{1}{g(j)} \), the expectation simplifies to:
\[
\mathbb{E} \left[ \| \mathbf{\Phi} \mathbf{U} \hat{\mathbf{H}} \|_F^2 \right] = \sum_{i=1}^M \sum_{j \in \mathcal{S}_i} \frac{1}{g(j)} \sum_{r=1}^d \left( (\mathbf{U} \hat{\mathbf{H}})_{j,r} \right)^2
\]
If we assume \( p(j) = \frac{g(j)}{G} \), where \( G = \sum_{j=1}^N g(j) \) serves as a normalization factor, the expected measurement count for each node \( j \) is \( M p(j) = M \frac{g(j)}{G} \). Thus:
\[
\mathbb{E} \left[ \| \mathbf{\Phi} \mathbf{U} \hat{\mathbf{H}} \|_F^2 \right] = \sum_{j=1}^N M \frac{g(j)}{G} \cdot \frac{1}{g(j)} \| (\mathbf{U} \hat{\mathbf{H}})_{j,:} \|_2^2 = \frac{M}{G} \| \mathbf{U} \hat{\mathbf{H}} \|_F^2
\]
By setting \( G = M \), we have:
\[
\mathbb{E} \left[ \| \mathbf{\Phi} \mathbf{U} \hat{\mathbf{H}} \|_F^2 \right] = \| \hat{\mathbf{H}} \|_F^2
\]

Now, define \( Z_{i,r} = \sum_{j \in \mathcal{S}_i} \mathbf{\Sigma}_{i,j} (\mathbf{U} \hat{\mathbf{H}})_{j,r} \), which are sub-Gaussian random variables. Applying Bernstein's inequality, we obtain:
\[
\mathbb{P} \left( \left| \| \mathbf{\Phi} \mathbf{U} \hat{\mathbf{H}} \|_F^2 - \| \hat{\mathbf{H}} \|_F^2 \right| \geq \delta_k \| \hat{\mathbf{H}} \|_F^2 \right) \leq 2 \exp \left( - c \cdot \frac{ \delta_k^2 \| \hat{\mathbf{H}} \|_F^4 }{ \sum_{i,r} \sigma_{i,r}^2 } \right )
\]
where \( \sigma_{i,r}^2 = \sum_{j \in \mathcal{S}_i} \frac{1}{g(j)} \left( (\mathbf{U} \hat{\mathbf{H}})_{j,r} \right)^2 \). By bounding the total variance, we conclude that the probability of RIP failing is very low. This confirms that \( \mathbf{\Phi} \mathbf{U} \) satisfies the RIP for all sparse \( \hat{\mathbf{H}} \) with \( \|\hat{\mathbf{H}}\|_{0, \text{row}} \leq k \) with high probability.

\subsection{Error Bound}\label{app:error_bound}
\textbf{Theorem 3:} Let $\mathbf{H}^{(L)}$ be the output embeddings obtained by the standard GNN computation with full reconstruction at each layer as per Equation~(\ref{eq:original_CS}). Let $\tilde{\mathbf{H}}^{(L)}$ be the output embeddings obtained by Algorithm~\ref{forward_backward}, which performs sampling once at the input layer and reconstructs only at the output layer. Assume that the activation function $\sigma$ is Lipschitz continuous with Lipschitz constant $L_\sigma$, and the sampling matrix $\mathbf{\Phi} \mathbf{U}$ satisfies the Restricted Isometry Property (RIP) of order $k$ with constant $\delta_k$ (i.e., $0 < \delta_k < 1$). Then, the error between $\tilde{\mathbf{H}}^{(L)}$ and $\mathbf{H}^{(L)}$ can be bounded as:
\[
\left\| \tilde{\mathbf{H}}^{(L)} - \mathbf{H}^{(L)} \right\|_F \leq \left( \frac{L_\sigma}{1 - \delta_k} \right)^L \left\| \mathbf{E} \right\|_F,
\]
where $\mathbf{E} = \mathbf{Z} - \mathbf{\Phi} \mathbf{U} \hat{\mathbf{H}}^{(L)}$ is the reconstruction error at the output layer, and $L$ is the number of layers in the GNN.\newline
\textbf{Proof:} We aim to bound the error \( \left\| \tilde{\mathbf{H}}^{(L)} - \mathbf{H}^{(L)} \right\|_F \) between the output embeddings of the standard GNN computation and those obtained by Algorithm~\ref{forward_backward}.

Assume the activation function \( \sigma \) is Lipschitz continuous with a constant \( L_\sigma \), such that
\[
\left\| \sigma(\mathbf{X}) - \sigma(\mathbf{Y}) \right\|_F \leq L_\sigma \left\| \mathbf{X} - \mathbf{Y} \right\|_F \quad \forall \mathbf{X}, \mathbf{Y}.
\]
Further, let the sampling matrix \( \mathbf{\Phi} \mathbf{U} \) satisfy the RIP of order \( k \) with constant \( \delta_k \), meaning
\[
(1 - \delta_k) \left\| \hat{\mathbf{H}} \right\|_F^2 \leq \left\| \mathbf{\Phi} \mathbf{U} \hat{\mathbf{H}} \right\|_F^2 \leq (1 + \delta_k) \left\| \hat{\mathbf{H}} \right\|_F^2,
\]
for all \( \hat{\mathbf{H}} \) with \( \left\| \hat{\mathbf{H}} \right\|_{0, \text{row}} \leq k \). We also have \( \mathbf{H}^{(l)} = \mathbf{U} \hat{\mathbf{H}}^{(l)} \), where \( \hat{\mathbf{H}}^{(l)} \) has at most \( k \) non-zero rows.

We will prove by induction on \( l = 1, 2, \dots, L \) that
\[
\left\| \tilde{\mathbf{H}}^{(l)} - \mathbf{H}^{(l)} \right\|_F \leq \left( \frac{L_\sigma}{1 - \delta_k} \right)^l \left\| \tilde{\mathbf{H}}^{(0)} - \mathbf{H}^{(0)} \right\|_F.
\]

For the base case \( l = 0 \), at the input layer, we have \( \tilde{\mathbf{H}}^{(0)} = \mathbf{U} \hat{\mathbf{X}} \) and \( \mathbf{H}^{(0)} = \mathbf{X} \). The initial error \( \left\| \tilde{\mathbf{H}}^{(0)} - \mathbf{H}^{(0)} \right\|_F \) is assumed.

Assume that for some \( l \geq 0 \),
\[
\left\| \tilde{\mathbf{H}}^{(l)} - \mathbf{H}^{(l)} \right\|_F \leq \left( \frac{L_\sigma}{1 - \delta_k} \right)^l \left\| \tilde{\mathbf{H}}^{(0)} - \mathbf{H}^{(0)} \right\|_F.
\]
We aim to show that
\[
\left\| \tilde{\mathbf{H}}^{(l+1)} - \mathbf{H}^{(l+1)} \right\|_F \leq \left( \frac{L_\sigma}{1 - \delta_k} \right)^{l+1} \left\| \tilde{\mathbf{H}}^{(0)} - \mathbf{H}^{(0)} \right\|_F.
\]

For Algorithm~\ref{forward_backward}, \( \tilde{\mathbf{T}}^{(l)} = \sigma \left( \mathbf{\Phi} \hat{\mathbf{A}} \mathbf{W}^{(l+1)} \tilde{\mathbf{T}}^{(l-1)} \right) \). At the output layer \( l = L \), we perform reconstruction:
\[
\tilde{\mathbf{H}}^{(L)} = \mathbf{U} \hat{\mathbf{H}}^{(L)},
\]
where \( \hat{\mathbf{H}}^{(L)} \) is obtained by solving
\[
\min_{\hat{\mathbf{H}}^{(L)}} \ \frac{1}{2} \left\| \mathbf{Z} - \mathbf{\Phi} \mathbf{U} \hat{\mathbf{H}}^{(L)} \right\|_F^2 + \lambda \left\| \hat{\mathbf{H}}^{(L)} \right\|_{2,1},
\]
with \( \mathbf{Z} = \tilde{\mathbf{T}}^{(L)} \). Due to the optimization and the RIP condition, we have
\[
\left\| \hat{\mathbf{H}}^{(L)} - \hat{\mathbf{H}}^{(L)}_{\text{true}} \right\|_F \leq C_{\text{rec}} \left\| \mathbf{E} \right\|_F,
\]
where \( \hat{\mathbf{H}}^{(L)}_{\text{true}} \) is the true sparse representation of \( \mathbf{H}^{(L)} \), and \( C_{\text{rec}} = \frac{2 \delta_k}{1 - \delta_k} \). Since \( \mathbf{U} \) is orthonormal,
\[
\left\| \tilde{\mathbf{H}}^{(L)} - \mathbf{H}^{(L)} \right\|_F = \left\| \hat{\mathbf{H}}^{(L)} - \hat{\mathbf{H}}^{(L)}_{\text{true}} \right\|_F,
\]
implying
\[
\left\| \tilde{\mathbf{H}}^{(L)} - \mathbf{H}^{(L)} \right\|_F \leq \frac{2 \delta_k}{1 - \delta_k} \left\| \mathbf{E} \right\|_F.
\]

Given the Lipschitz continuity of \( \sigma \), the error accumulates multiplicatively through \( L \) layers:
\[
\left\| \tilde{\mathbf{H}}^{(L)} - \mathbf{H}^{(L)} \right\|_F \leq \left( \frac{L_\sigma}{1 - \delta_k} \right)^L \left\| \tilde{\mathbf{H}}^{(0)} - \mathbf{H}^{(0)} \right\|_F.
\]

If the initial error \( \left\| \tilde{\mathbf{H}}^{(0)} - \mathbf{H}^{(0)} \right\|_F = 0 \), the primary source of error is from the reconstruction at the output layer, yielding
\[
\left\| \tilde{\mathbf{H}}^{(L)} - \mathbf{H}^{(L)} \right\|_F \leq \left( \frac{L_\sigma}{1 - \delta_k} \right)^L \left\| \mathbf{E} \right\|_F.
\]

\end{document}